\documentclass{article} 
\usepackage{collas2024_conference,times}
\usepackage{easyReview}

\usepackage{amsmath,amsfonts,bm}

\def\eqref#1{equation~\ref{#1}}

\def\1{\bm{1}}

\DeclareMathAlphabet{\mathsfit}{\encodingdefault}{\sfdefault}{m}{sl}
\SetMathAlphabet{\mathsfit}{bold}{\encodingdefault}{\sfdefault}{bx}{n}

\usepackage{hyperref}
\hypersetup{
    colorlinks=true,
    linkcolor=red,
    filecolor=magenta,
    urlcolor=blue,
    citecolor=purple,
    pdftitle={Overleaf Example},
    pdfpagemode=FullScreen,
    }

\usepackage{algorithm}
\usepackage[noend]{algpseudocode}
\usepackage{amsmath}
\usepackage{amssymb}
\usepackage{mathtools}
\usepackage{amsthm}
\usepackage{bm}
\theoremstyle{plain}

\usepackage{wrapfig}
\usepackage{caption}
\usepackage{subcaption}
\usepackage{multirow}
\usepackage{xcolor}

\newcommand{\planet}{PlaNet }
\newcommand{\dreamer}{Dreamer }
\newcommand{\loca}{LoCA }
\newcommand{\mcloca}{MountainCarLoCA }
\newcommand{\mgloca}{MiniGridLoCA }
\newcommand{\reachloca}{ReacherLoCA }
\newcommand{\rndreachloca}{RandomReacherLoCA }

\title{Partial Models for Building Adaptive Model-Based Reinforcement Learning Agents}

\author{
Safa Alver$^{1,3}$, Ali Rahimi-Kalahroudi$^{2,3}$, Doina Precup$^{1,3,4}$\\
$^{1}$McGill University, $^{2}$University of Montreal, $^{3}$Mila Quebec AI Institute, $^{4}$Google DeepMind\\
\texttt{\small safa.alver@mail.mcgill.ca}, \texttt{\small ali-rahimi.kalahroudi@mila.quebec},
\texttt{\small dprecup@cs.mcgill.ca} \\
}

\collasfinalcopy 

\begin{document}

\maketitle

\begin{abstract}
In neuroscience, one of the key behavioral tests for determining whether a subject of study exhibits model-based behavior is to study its adaptiveness to local changes in the environment. In reinforcement learning, however, recent studies have shown that modern model-based agents display poor adaptivity to such changes. The main reason for this is that modern agents are typically designed to improve sample efficiency in single task settings and thus do not take into account the challenges that can arise in other settings. In local adaptation settings, one particularly important challenge is in quickly building and maintaining a sufficiently accurate model after a local change. This is challenging for deep model-based agents as their models and replay buffers are monolithic structures lacking distribution shift handling capabilities. In this study, we show that the conceptually simple idea of \emph{partial models} can allow deep model-based agents to overcome this challenge and thus allow for building locally adaptive model-based agents. By modeling the different parts of the state space through different models, the agent can not only maintain a model that is accurate across the state space, but it can also quickly adapt it in the presence of a local change in the environment. We demonstrate this by showing that the use of partial models in agents such as deep Dyna-Q, PlaNet and Dreamer can allow for them to effectively adapt to the local changes in their environments.
\end{abstract}

\section{Introduction}
\label{sec:intro}

Recent studies have shown that modern model-based reinforcement learning (MBRL) agents display poor signs of adaptivity to the local changes in their environments \citep{van2020loca, wan2022towards}, despite this being a key behavioral characteristic of model-based biological agents \citep{daw2011model}. The analysis of \citet{wan2022towards} reveals that the main reason for this lack of adaptivity is because of the agent's inability in building and maintaining a sufficiently accurate model after a local change. This is a challenge for modern model-based agents as their models and replay buffers are monolithic structures lacking distribution shift handling capabilities. More specifically, in deep model-based agents the data that is used in updating the agent's model is stored in a single replay buffer and thus, in the face of a local change, leads to problems like (i) the interference of the old and new data and (ii) the forgetting of the old-but-relevant data \citep{wan2022towards, rahimi2023replay}. Moreover, the data that is used in the updates is sampled in a random fashion, which leads to problems like the agent ending up with a biased model. Finally, the model that is used in updating the agent's policy is a single model and thus leads to problem like the inability to perform to-the-point updates for quick adaptation in the face of a local change.

To address these challenges, we propose the use of \emph{partial models} \citep[see e.g.,][]{talvitie2008simple, khetarpal2021temporally, zhao2021consciousness, alver2023minimal}. Under this scenario, in its simplest implementation, the agent first detects the number of required partial models and then maintains a separate model for each relevant part of the state space, and in each update step before the local change, it performs updates to all of its models. Then, after the local change, it only updates the necessary models. And, in its scalable implementation, the agent emulates the idea of maintaining separate models by using separate heads and index lists. This conceptually simple idea naturally leads to a model, which is a collection of multiple partial models, that is both accurate across the state space and also quickly adaptable in the face of a local change in the environment.

We demonstrate the effectiveness of partial models by first instantiating them in the deep Dyna-Q agent and showing that they allow for achieving local adaptivity on the Local Change Adaptation (LoCA) setup \citep{van2020loca, wan2022towards} of both the MountainCar and MiniGrid domains. We then test the generality of these results by instantiating partial models in modern deep MBRL agents like \planet \citep{hafner2019learning} and \dreamer \citep{Hafner2020Dream, hafner2021mastering, hafner2023mastering}. Experiments on the \loca setups of the pixel-based MuJoCo Reacher and RandomReacher domains demonstrate that the use of partial models can indeed drastically improve the local adaptation capability of modern MBRL agents as well.

\textbf{Key Contributions.} The key contributions of this study are as follows: (i) Compared to the previous studies on local adaptation \citep{van2020loca, wan2022towards, rahimi2023replay}, we propose two additional versions of the \loca setup, with stochastic reward functions and non-stationary transition distributions, that are both more challenging than the original one (Sec.\ \ref{sec:add_loca_setups}). (ii) We provide a detailed discussion on the challenges that arise in building locally adaptive deep MBRL agents and identify two additional challenges (Sec.\ \ref{sec:challenges}). (iii) We propose two instantiations of the idea of partial models, a simple and scalable one, and demonstrate across four different domains that these types of models can allow for building locally adaptable deep MBRL agents (Sec.\ \ref{sec:partial_models}, \ref{sec:nonlinear_dyna} \& \ref{sec:dreamer_planet}).

\section{Background}
\label{sec:background}

\textbf{Reinforcement Learning.} In RL \citep{sutton2018reinforcement}, an agent interacts with its environment through a sequence of actions to maximize its long-term cumulative reward. The interaction is usually modeled as a Markov decision process (MDP) $(\mathcal{S}, \mathcal{A}, P, R, \rho_0, \gamma)$, where $\mathcal{S}$ and $\mathcal{A}$ are the (finite) set of state and actions, $P:\mathcal{S}\times\mathcal{A}\to \text{Dist}(\mathcal{S})$ is the transition distribution, $R: \mathcal{S}\times\mathcal{A}\times\mathcal{S}\to \mathbb{R}$ is the reward function, $\rho_0: \mathcal{S}\to\text{Dist}(\mathcal{S})$ is the initial state distribution, and $\gamma\in [0, 1)$ is the discount factor. The goal of the agent is to obtain a policy $\pi: \mathcal{S}\to \text{Dist}(\mathcal{A})$ that maximizes the expected sum of discounted rewards $E_{\pi} [\sum_{t=0}^{\infty} \gamma^t R(S_t, A_t, S_{t+1}) | S_0 \sim \rho_0]$.

\textbf{Deep Model-Based Reinforcement Learning.} In deep MBRL, an agent obtains a policy by planning with a learned model $m$ of the environment which is represented with deep neural networks. Even though various deep MBRL agents have been proposed in the recent years \citep{moerland2023model}, notable state-of-the-art examples of them are \planet \citep{hafner2019learning} and \dreamer \citep{Hafner2020Dream, hafner2021mastering, hafner2023mastering}, which perform decision-time and background planning, respectively \citep{sutton2018reinforcement, alver2022understanding}. These agents use deep neural networks in the implementation of their models and value functions, and store their experiences into a replay buffer. The stored experiences are then used for learning a model of the environment, which is then used for updating the value function of the agent.

\begin{wrapfigure}{R}{0.5\textwidth}
    \vspace{-0.25cm}
    \centering
    \includegraphics[width=7.5cm]{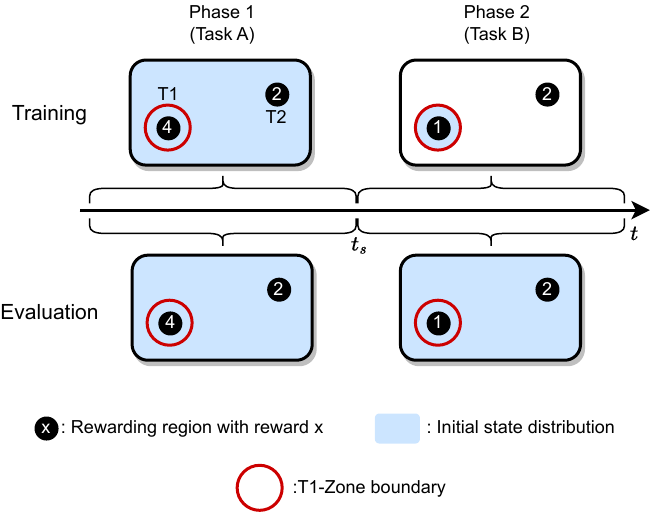}
    \caption{The LoCA setup of \citet{wan2022towards}. The values in the rewarding regions indicate the reward that is received in the corresponding region. $t_s$ indicates the point in which the phase shift happens.
    }
    \vspace{-0.25cm}
    \label{fig:loca_setup}
\end{wrapfigure}

\textbf{The LoCA Setup.} Inspired by studies on detecting model-based behavior in biological agents \citep{daw2011model}, \citet{van2020loca} proposed the LoCA setup to evaluate model-based behavior in RL agents. Later, \citet{wan2022towards} improved this setup by making it (i) simpler, (ii) less sensitive to hyperparameters and (iii) easily applicable to stochastic environments. The LoCA setup measures the adaptivity of an agent and thus serves as a preliminary yet important step towards the continual RL problem \citep{khetarpal2022towards}.

The LoCA setup considers an environment with two tasks, namely Task A and Task B, which differ only in their reward functions (see the first and second columns of Fig.\ \ref{fig:loca_setup}, respectively). In each task, there are two rewarding regions, namely T1 and T2, and the reward function for the states outside of this region is always 0. For task A, the agent receives a reward of +4 upon entering T1 and +2 upon entering T2. For task B, however, entering T1 yields a reward of +1 (instead of +4 as in task A), and entering T2 yields the same reward as in task A, which is +2. Finally, the transition dynamics for a local area around T1 (called the T1-zone) is such that it is impossible to get out, once entered. Note that the difference between the two tasks is local and is only in the reward functions. Also note that while the difference is only local, the optimal policies are completely different: while the optimal policy for task A points to T1, the optimal policy for task B points to T2 (except when the agent is within the T1-zone, where the optimal policy again points to T1).

To test for adaptivity, the LoCA setup considers a scenario with two consecutive training phases, namely Phase 1 and Phase 2, which differ in the tasks and initial state distributions (see the first row of Fig.\ \ref{fig:loca_setup}). Throughout Phase 1, the task is task A and the initial state is drawn uniformly from the entire state space. After Phase 1, Phase 2 begins and the task switches to task B (see point $t_s$ in Fig.\ \ref{fig:loca_setup}). Now, the initial state is drawn uniformly from only states within the T1-zone; hence, the agent is stuck in the local area around T1 in Phase 2. Note that the agent does not receive any signal upon a phase switch. Throughout both of the training phases, the agent is periodically evaluated and the initial state for each evaluation episode is drawn uniformly over the entire state space (see the second row of Fig.\ \ref{fig:loca_setup}). In each evaluation interval, the mean return is calculated and it is compared to the mean return of the corresponding optimal policy in that phase.

For the agent to reach optimal performance in Phase 2, it has to adapt to the new reward function that is present in task B: it has to change its policy from pointing towards T1 to pointing towards T2 for most of the states in the state space (states outside the T1-zone). Note that the agent has to perform this adaptation while only observing states within the T1-zone. Given a sufficient amount of time to train in both phases, the LoCA setup classifies an agent as adaptive if it is able to achieve close-to-optimal performance in both Phase 1 and Phase 2. If the agent only achieves near-optimal performance in Phase 1 but not in Phase 2, it is classified as non-adaptive. And lastly, the LoCA setup makes no assessment if the agent fails to reach close-to-optimal performance in Phase 1.

\section{More Challenging Local Adaptation Setups}
\label{sec:add_loca_setups}

\begin{table}
    \centering
    \caption{The details of the reward functions and transition distributions of LoCA, LoCA1 and LoCA2 setups.}
        \begin{tabular}{|l|c|c|c|c|} 
        \cline{2-5}
        \multicolumn{1}{l|}{}  & \multicolumn{2}{c|}{Reward Function} & \multicolumn{2}{c|}{Transition Distribution}                                                                                                              \\ 
        \cline{2-5}
        \multicolumn{1}{l|}{}  & Phase 1        & Phase 2             & Phase 1                         & Phase 2                                                                                                                 \\ 
        \hline
        \multirow{4}{*}{LoCA}  & \textcolor{red}{D}eterministic & \textcolor{red}{D}eterministic      & \multirow{4}{*}{\textcolor{red}{D}eterministic} & \multirow{4}{*}{\textcolor{red}{D}eterministic}                                                                                         \\
                               & T1 region: +4  & T1 region: +1       &                                 &                                                                                                                         \\
                               & T2 region: +2  & T2 region: +2       &                                 &                                                                                                                         \\
                               & Rest: 0        & Rest: 0             &                                 &                                                                                                                         \\ 
        \hline
        \multirow{4}{*}{LoCA1} & \textcolor{blue}{S}tochastic     & \textcolor{blue}{S}tochastic          & \multirow{4}{*}{\textcolor{red}{D}eterministic} & \multirow{4}{*}{\textcolor{red}{D}eterministic}                                                                                         \\
                               & T1 region: $\mathcal{N}(+4, 0.5)$  & T1 region: $\mathcal{N}(+1, 0.5)$       &                                 &                                                                                                                         \\
                               & T2 region: $\mathcal{N}(+2, 0.5)$  & T2 region: $\mathcal{N}(+2, 0.5)$       &                                 &                                                                                                                         \\
                               & Rest: $\mathcal{N}(0, 0.5)$       & Rest: $\mathcal{N}(0, 0.5)$            &                                 &                                                                                                                         \\ 
        \hline
        \multirow{4}{*}{LoCA2} & \textcolor{blue}{S}tochastic     & \textcolor{blue}{S}tochastic          & \multirow{4}{*}{\textcolor{red}{D}eterministic} & \multirow{4}{*}{\begin{tabular}[c]{@{}c@{}}\textcolor{blue}{S}tochastic \\(Slippery T1-zone \\with slip probability 0.2)\end{tabular}}  \\
                               & T1 region: $\mathcal{N}(+4, 0.5)$  & T1 region: $\mathcal{N}(+1, 0.5)$       &                                 &                                                                                                                         \\
                               & T2 region: $\mathcal{N}(+2, 0.5)$  & T2 region: $\mathcal{N}(+2, 0.5)$       &                                 &                                                                                                                         \\
                               & Rest: $\mathcal{N}(0, 0.5)$       & Rest: $\mathcal{N}(0, 0.5)$            &                                 &                                                                                                                         \\
        \hline
        \end{tabular}
    \label{tab:loca_table}
\end{table}

Even though the LoCA setup provides a way to evaluate model-based behavior in RL agents, it has two main shortcomings: it only considers (i) scenarios with deterministic reward functions, and (ii) scenarios with stationary transition distributions (see the first row of Table \ref{tab:loca_table}). In order to overcome these shortcomings, we propose two additional versions of the LoCA setup that are both more challenging that the original version: (i) LoCA1: a version of the setup with a stochastic reward function, and (ii) LoCA2: a version of the setup with both a stochastic reward function and a non-stationary transition distribution. More specifically, we convert the reward function into a stochastic one by turning it into a Gaussian: in Phase 1, upon entering T1 and T2 the agent receives a reward that is sampled from a Gaussian distribution with mean +4 and +2, respectively. And, we introduce non-stationarity into the transition dynamics by making the T1-zone slippery in Phase 2. More details on these setups can be found in the last two rows of Table \ref{tab:loca_table}.

\section{Challenges in Building Adaptive Deep Model-Based Agents}
\label{sec:challenges}

\textbf{Reason for Failing to Adapt.} Previous studies on local adaptation \citep{van2020loca, wan2022towards} have reported that popular deep MBRL agents, such as PlaNet \citep{hafner2019learning} and Dreamer \citep{Hafner2020Dream, hafner2021mastering, hafner2023mastering}, fail in adapting to the local changes in their environments. The analysis of \citet{wan2022towards} reveals that the key reason for the failure is because of the agent's inability in building and maintaining a sufficiently accurate model after a phase shift happens in the \loca setup. 

\textbf{Challenge 1: Interference-Forgetting Dilemma.} As pointed out in \citet{wan2022towards} and \citet{rahimi2023replay}, one of the key challenges in building adaptive deep model-based agents is to overcome the \emph{interference-forgetting dilemma}, which prevents an agents from building and maintaining a sufficiently accurate model after the phase change. In this dilemma, depending on the size of its replay buffer, the agent either faces the problem of interference or forgetting. More specifically, in the case of having a large replay buffer the agent faces the problem of \emph{interference} which is caused by the interference of the stale data in the buffer with the new incoming data: e.g. in the \loca setup, after a phase shift happens the agent will be receiving +1 rewarding transitions, however, its replay buffer will still be containing the stale +4 rewarding transitions from Phase 1 and because of this the updates to the model corresponding to the T1 region will be interfered with the stale transitions. This will then result in an inaccurate model, which will then affect planning and thereby render the agent non-adaptive. And, in the case of using a small replay buffer the agent faces the problem of \emph{forgetting} which is caused by the loss of the old-but-relevant data in the buffer: e.g. in the \loca setup, after a phase shift happens the agent will be receiving +1 rewarding transitions and, due to the limited size of its buffer, it will over time lose the old-but-relevant +2 rewarding transitions from Phase 1 and because of this the model will over time fail in generating transitions with a reward of +2. This will again result in an inaccurate model, which will then again affect planning and thereby render the agent non-adaptive.

\textbf{Challenge 2: Proper Model Update Challenge.} Even if the agent manages to overcome the interference-forgetting dilemma, another very important challenge is in performing proper updates to the model so that it becomes and remains sufficiently accurate after a phase change. As modern deep MBRL agents update their models with randomly-sampled batches of transitions, they face the problem of ending up with biased models that are skewed towards the abundant transitions in the replay buffer: e.g. in the \loca setup, after a phase shift happens the agent will be receiving lots of +1 rewarding transitions and because of this its model will be skewed towards generating transitions with a reward of +1. This problem will again result in an inaccurate model, which will then again affect planning and thereby render the agent non-adaptive. We refer to this challenge as the \emph{proper model update challenge}.

\textbf{Challenge 3: Quick Adaptation Challenge.} Lastly, even though the previously introduced challenges are vital challenges in on their own, building agents that are not only adaptable but also \emph{quickly} adaptable is another major challenge, as quick adaptation is another key characteristic of model-based behavior \citep{daw2011model}. We refer to this challenge as the \emph{quick adaptation challenge}.

\section{Partial Models for Building Adaptive Model-Based Agents}
\label{sec:partial_models}

Before introducing the idea of partial models, we first introduce the terminology that we will use for the models and replay buffers of deep MBRL agents: following \citet{van2019use}, we refer to their models and replay buffers as \emph{parametric} and \emph{non-parametric} models , respectively.\footnote{While the former terminology is already obvious, the latter one also becomes obvious if one views the replay buffer as a model that provides accurate transitional data from the agent's past experiences.} We choose to use this terminology as it allows for a coherent presentation of the idea of partial models.

\subsection{The Simplest Implementation}
\label{sec:simple_imp}

\begin{wrapfigure}{R}{0.4\textwidth}
    \centering
    \includegraphics[width=0.35\textwidth]{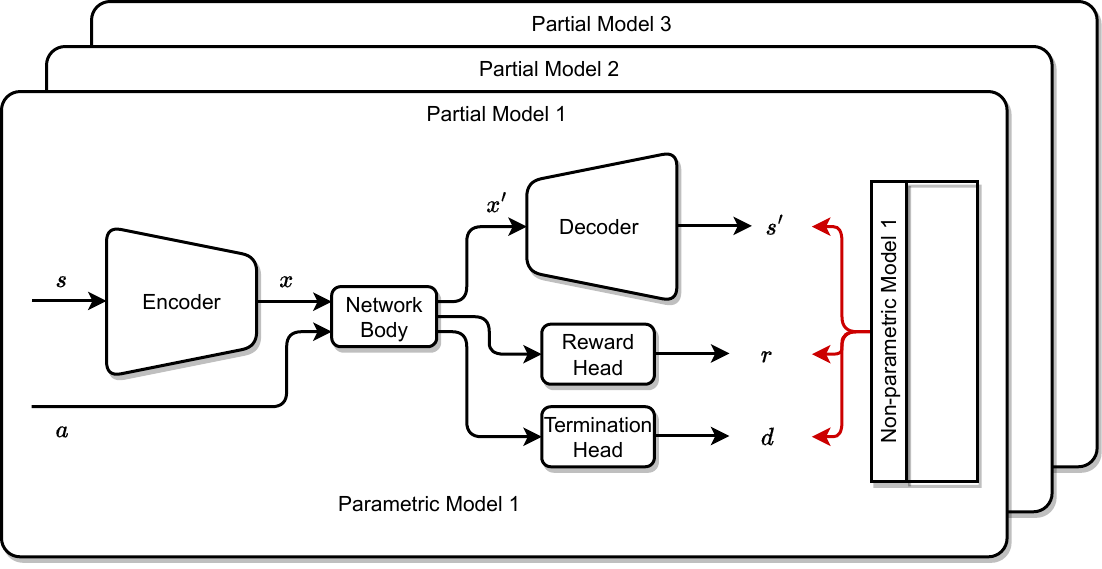}
    \caption{An instantiation of the simplest implementation of partial models with three pairs of non-parametric and parametric models. The red arrows indicate the direction of information flow from the non-parametric models to the parametric ones. The parametric model consist of an encoder $e$, network body $nb$, decoder $d$, reward head $rh$ and termination head $th$.}
    \vspace{-0.5cm}
    \label{fig:mult_model_arch}
\end{wrapfigure}

\textbf{Details of the Architecture.} In this study, we propose to use \emph{partial models} as a way to address all of the three challenges presented in Sec. \ref{sec:challenges}. In its simplest implementation, instead of having a single pair of non-parametric and parametric models, as is the default in deep MBRL agents, the agent maintains multiple pairs of non-parametric and parametric models that are each responsible for modeling different parts of the state space. An instantiation of this implementation with three partial models is illustrated in Fig.\ \ref{fig:mult_model_arch}. 

Two important questions that arise in this context are (i) how many partial models to use, and (ii) how to determine which parts of the state space would each partial model be modeling. In the \loca setup, as rewards allow for a natural way to cluster the state space, the agent uses of them to provide answers to these questions. More specifically, by using the data that is collected in the initial exploration phase, it first learns a neural embedding function $E$ of the states such that states that have a similar reward are also closer in their embedding representation using contrastive learning \citep{hadsell2006dimensionality, dosovitskiy2014discriminative}, and then runs a clustering algorithm, e.g. k-means clustering, over the embeddings and rewards to determine the number of partial models. Finally, it assigns each of the $n$ models to parts of the state space that belongs to a different cluster.

\begin{wrapfigure}{R}{0.6\textwidth}
\vspace{-0.75cm}
\centering
\begin{minipage}{1.0\linewidth}
    \begin{algorithm}[H]
        \centering
        \caption{Pseudocode for achieving adaptivity with the Simple and Scalable Implementation of Partial Models. The \textcolor{red}{red} and \textcolor{blue}{blue} colored parts are specific to the \textcolor{red}{simple} and \textcolor{blue}{scalable} implementations, respectively (i.e. the simple implementation does not contain the blue parts and the scalable implementation does not contain the red parts).}\label{alg:brief_pseudocode}
        \footnotesize
        \begin{algorithmic}[1]

        \State $\mathcal{D} \gets$ gather $m$ samples via a random policy
        \State $\mathcal{CC}$, $C$, $E$, $n \gets $ \textsc{IdentifyClusters\&ObtainClassifier}($\mathcal{D}$) (\textsc{ICOC}; $\mathcal{CC}$ is a dictionary of cluster centers, $C$ is the classifier, $E$ is the neural embedding function and $n$ is the number of clusters)
        
        \State {\color{red} $\mathcal{NP}\gets \{ m_{np}^i = [\text{ }] \}_{i=1}^n$ (initialize $n$ empty non-parametric models)}
        \State {\color{red} $\mathcal{P}\gets  \{ m_{p}^i = (e, nb, d, rh, th) \}_{i=1}^n$ (initialize $n$ parametric models)}
        
        \State {\color{blue} $m_{np} \gets [\text{ }]$ (initialize an empty non-parametric model)}
        \State {\color{blue} $\mathcal{IL}\gets \{\mathcal{I}^i = [\text{ }]\}_{i=1}^n$ (initialize $n$ empty index lists)}
        \State {\color{blue} $m_p \gets (e, nb, d, \{ rh^i\}_{i=1}^{n}, th)$ (initialize a parametric model with $n$ $rh$'s)}
        \State Initialize the parameters of the parametric models and of $Q$
        \State
        
        \State $phase\gets 1 \text{ (the current LoCA phase)}$

        \While{training continues}
        \State $S\gets \text{reset environment}$
        \While{\text{not done}}
        \State \text{$R, S', \text{done} \gets \text{environment($\epsilon\text{-greedy}(Q(S))$)}$}

        \If{$C(\text{concatenate}(E(S), R)$ = ``anomalous''}
            \State $phase\gets 2$
            \State $k_{\text{req\_upd}} \gets $ detect which model requires update by identifying \Statex \hspace*{30mm}the cluster ID that $E(S)$ is closest using $\mathcal{CC}$
            \State Clear {\color{red} $\mathcal{NP}[k_{\text{req\_upd}}]$} / {\color{blue} $\mathcal{IL}[k_{\text{req\_upd}}]$}
            \State $\mathcal{D} \gets$ gather $m$ samples via a random policy 
            \State $\mathcal{CC}_{\text{new}}$, $C_{\text{new}}$, $E$, $\_ \gets $ \textsc{ICOC}({\color{red} $\mathcal{D} + \mathcal{NP}$} / {\color{blue} $\mathcal{D}$ + $m_{np}$ + $\mathcal{IL}$})
            \State $\mathcal{CC}$, $C \gets $ compare $\mathcal{CC}$ and $\mathcal{CC}_{\text{new}}$, identify similar clusters \Statex \hspace*{28mm}using the cluster centers, and retain the old cluster \Statex \hspace*{28mm}IDs of these similar ones in $\mathcal{CC}_{\text{new}}$ and $C_{\text{new}}$
        \EndIf

        \State $k\gets $ identify the model class by $C(\text{concatenate}(E(S), R)$
        \State {\color{red} $\mathcal{NP}[k]$} / {\color{blue} $m_{np}$} $\gets$ {\color{red} $\mathcal{NP}[k]$} / {\color{blue} $m_{np}$} $+ \{(S,A,R,S', \text{done})\}$
        \State {\color{blue} \text{$\mathcal{IL}[k]\gets \mathcal{IL}[k] + \{ \text{current time step } t \} $}}
        
        \If{$phase = 1$}
        \For{$i \text{ in } \{1...n\}$}
            \State $\mathcal{B}\gets \text{sample\_batch}({\color{red}\mathcal{NP}[i]} / {\color{blue} m_{np}, \mathcal{IL}[i]})$
            \State Update {\color{red} $\mathcal{P}[i]$} / {\color{blue} $m_p[i]$} with $\mathcal{B}$
            \State Update $Q$ with the predictions of {\color{red} $\mathcal{P}[i]$} / {\color{blue} $m_p[i]$} on $\mathcal{B}$
        \EndFor
        \EndIf

        \If{$phase = 2$}
        \State $\mathcal{B}\gets \text{sample\_batch}({\color{red} \mathcal{NP}[k_{\text{req\_upd}}]} / {\color{blue} m_{np}, \mathcal{IL}[k_{\text{req\_upd}}} )$
        \State {\color{red} Update $\mathcal{P}[k_{\text{req\_upd}}]$ with $\mathcal{B}$}
        \State {\color{blue} Freeze $(e, nb, d, th)$ in $m_p$ and only update $rh^{k_{\text{req\_upd}}}$ with $\mathcal{B}$}
        \For{$i \text{ in } \{1...n\}$}
            \State $\mathcal{B}\gets \text{sample\_batch}({\color{red}\mathcal{NP}[i]} / {\color{blue} m_{np}, \mathcal{IL}[i]})$
            \State Update $Q$ with the predictions of {\color{red} $\mathcal{P}[i]$} / {\color{blue} $m_p[i]$} on $\mathcal{B}$
        \EndFor
        \EndIf
        
        \State $S\gets S'$
        \EndWhile
        \EndWhile
        \end{algorithmic}
    \end{algorithm}
\end{minipage}
\vspace{-0.75cm}
\end{wrapfigure}

In the learning of the neural embedding function, similar to \citet{rahimi2023replay}, we learn an embedding function $v = E(S)$, where $E$ is a deep neural network that maps state $S$ to an embedding vector $v$. This embedding function induces a distance metric in the state-space, such as $d(S_i, S_j) = ||E(S_i) - E(S_j)||_2$ for states $S_i$ and $S_j$. We train the embedding function such that states that have similar rewards, have a smaller distance between the learned embeddings and states that have dissimilar rewards, have a relatively larger distance between them in the embedding space. Let $S'$ represent a state that is in the reward proximity of state $S$, i.e.\ $|r(S) - r(S')| < p$ for some $p$, and $\bar{\mathbf{S}}$ represent a set of states that are not, and let $D = \{S, S', \bar{\mathbf{S}}\}$ be a dataset of their collection. We train the embedding function to minimize the following loss function:
\begin{align*}
    \textstyle
    L(D) = \sum_{(S,S',\bar{\mathbf{S}}) \in D} ||E(S) - E(S')||_2^2 \nonumber + \left(\beta - \sum_{\bar{S}\in \bar{\mathbf{S}}} ||E(S) - E(\bar{S})||_2^2 \right)^2,
\end{align*}
where $\beta > 0$ is a hyperparameter. This loss function trains the embedding of states $S$ and $S'$ to be closer by minimizing their distance towards zero, while pushing the embedding of states $\bar{\mathbf{S}}$ to be on average farther, with a cumulative squared distance value close to $\beta$.

\textbf{Details of Achieving Adaptivity.} Once these issues are addressed, the agent would be able to use partial models to achieve local adaptivity. Under this scenario, in Phase 1, while performing updates to its parametric models $\mathcal{P}=\{m_{p}^i = (e, nb, d, rh, th) \}_{i=1}^n$ and value function $Q$, the agent updates (i) all of its $n$ parametric models with their corresponding non-parametric models $\mathcal{NP}=\{m_{np}^i\}_{i=1}^n$ and (ii) its value function with all of its parametric models in $\mathcal{P}$. And, in Phase 2, it first clears the stale samples in its corresponding non-parametric model and then updates (i) only the necessary parametric model with its corresponding non-parametric one and (ii) its value function with all of its parametric models.

Note that in order to only perform the necessary updates in Phase 2, the agent would have to first (i) detect the phase change and then (ii) infer which of its parametric models requires an update, as the LoCA setup does not provide this information. To overcome these issues, the agent trains a classifier $C$ to distinguish between the rewards it has seen so far and the anomalous ones, and after detecting the change, it identifies the model that requires an update by first passing the state corresponding to the anomalous reward through the embedding function and then identifying which cluster it belongs to. The main pseudocode of how to achieve adaptivity with the simple implementation of partial models is presented in Alg.\ \ref{alg:brief_pseudocode}. For more details on the \textsc{ICOC} algorithm, we refer the reader to Alg.\ \ref{alg:icoc_alg} in App.\ \ref{app:icoc}.

\textbf{How does this implementation address the challenges?} Having modularity in the non-parametric model structure allows the agent to mitigate the interference-forgetting dilemma as the agent now clears the stale samples in its corresponding non-parametric model upon a detection of a phase change, and it also stores the old-but-relevant data in a separate non-parametric model. This modularity also addresses the proper model update challenge as the agent now samples an equal amount of transitions from each of its non-parametric models to update its corresponding parametric ones. Lastly, the modularity in the parametric model structure also addresses the quick adaptation challenge as the agent now infers which of its parametric models requires an update upon a phase change and it focuses on updating the one that requires it. 

\subsection{A Scalable Implementation}
\label{sec:scalable_imp}

\begin{wrapfigure}{R}{0.4\textwidth}
    \centering
    \includegraphics[width=0.35\textwidth]{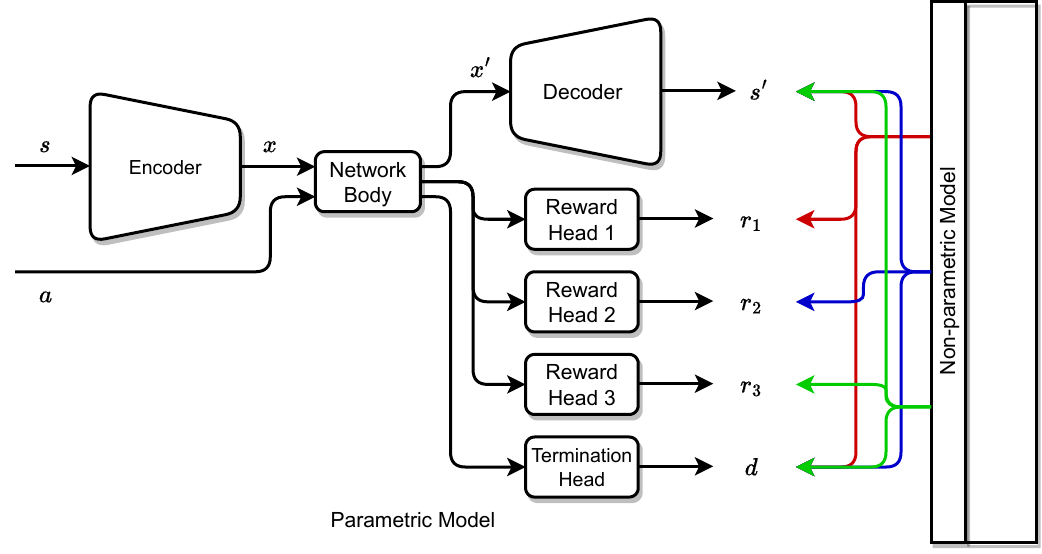}
    \caption{An instantiation of a scalable implementation of partial models with a non-parametric model consisting of three index lists and a parametric one consisting of three reward heads. The red, blue and green arrows indicate the direction of information flow from the non-parametric model to the parametric one.}
    \vspace{-0.5cm}
    \label{fig:one_model_arch}
\end{wrapfigure}

Even though the simple implementation allows for building adaptive MBRL agents, maintaining multiple models is not a scalable approach as the memory and computation requirements grow linearly with each model. Thus, we now present a scalable implementation of the idea of partial models that again retains simplicity. 

\textbf{Details of the Architecture.} In this implementation, the agent now maintains a single non-parametric and parametric model that emulates the idea of having multiple models: after determining how many partial models to use and which parts of the state space would each partial model be modeling through the same procedure as in Sec.\ \ref{sec:simple_imp}, the non-parametric model $m_{np}$ now maintains a separate index list $\mathcal{IL} = \{\mathcal{I}^i \}_{i=1}^n$ for each of the partial models, and the parametric model $m_p = (e, nb, d, \{ rh^i\}_{i=1}^{n}, th)$ now consists of multiple reward heads each corresponding to a different partial model. An instantiation of this implementation with three partial models is illustrated in Fig.\ \ref{fig:one_model_arch}.

\textbf{Details of Achieving Adaptivity.} Under this scenario, in Phase 1, while performing updates to its parametric model $m_p$, the agent (i) first samples an equal amount of transitions by using each index list $\mathcal{I}^i$ in its non-parametric model $m_{np}$ and (ii) then it updates its encoder $e$, network body $nb$, decoder $d$ and termination head $th$ with all of these transitions. However, while updating its reward heads $rh^i$, it only updates them with the transitions that are sampled from their corresponding index lists. And, in Phase 2, the agent (i) first clears the list that corresponds to the indices of the stale samples, and (ii) then freezes all the parameters of its parametric model except for the parameters of the reward head that requires adaptation. Then, it updates only these parameters with the transitions that are sampled according to the corresponding index list. Note that, in both of the phases, while performing updates to its value function $Q$, the agent makes use of the transitions that are generated from all of the heads of its parametric model. Finally, (i) the detection of the phase change and (ii) the inference of which reward head to perform updates to is done in a similar fashion as in Sec. \ref{sec:simple_imp}. The main pseudocode of how to achieve adaptivity with the scalable implementation of partial models is presented in Alg.\ \ref{alg:brief_pseudocode}.

\textbf{How does this implementation address the challenges?} Note that this implementation also addresses both the interference-forgetting dilemma and the proper model update challenge as the agent again maintains modularity in the non-parametric model structure through the separate index lists. And, it also addresses the quick adaptation challenge as the agent again maintains modularity in its parametric model through its separate reward heads.

\section{Deep Dyna-Q Experiments}
\label{sec:nonlinear_dyna}

Following previous studies on local adaptation \citep{wan2022towards, rahimi2023replay}, we start by performing experiments with the deep Dyna-Q agent as it is a simple MBRL agent that is reflective of many of the properties of its state-of-the-art counterparts \citep{hafner2019learning, Hafner2020Dream, hafner2021mastering, hafner2023mastering}. The details of all of the experiments in this section can be found in App.\ \ref{app:mountaincar_appendix} \& \ref{app:minigrid_appendix}.

\begin{wrapfigure}{R}{0.5\textwidth}
    \centering
    \begin{subfigure}{0.23\textwidth}
        \centering
        \includegraphics[width=3.5cm]{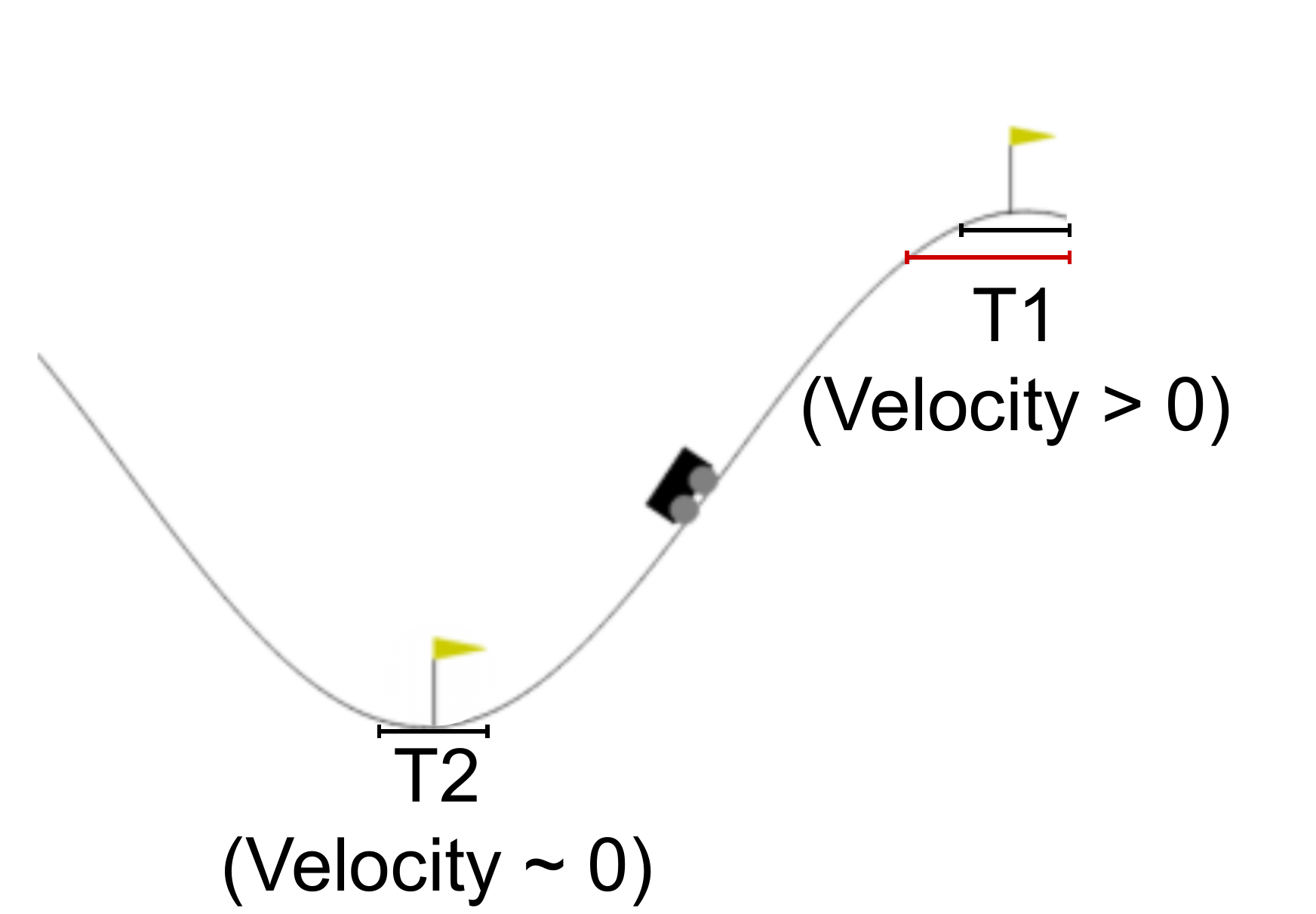}
        \caption{MountainCarLoCA} \label{fig:mountaincar_illus}
    \end{subfigure}
    \begin{subfigure}{0.23\textwidth}
        \centering
        \includegraphics[width=2.25cm]{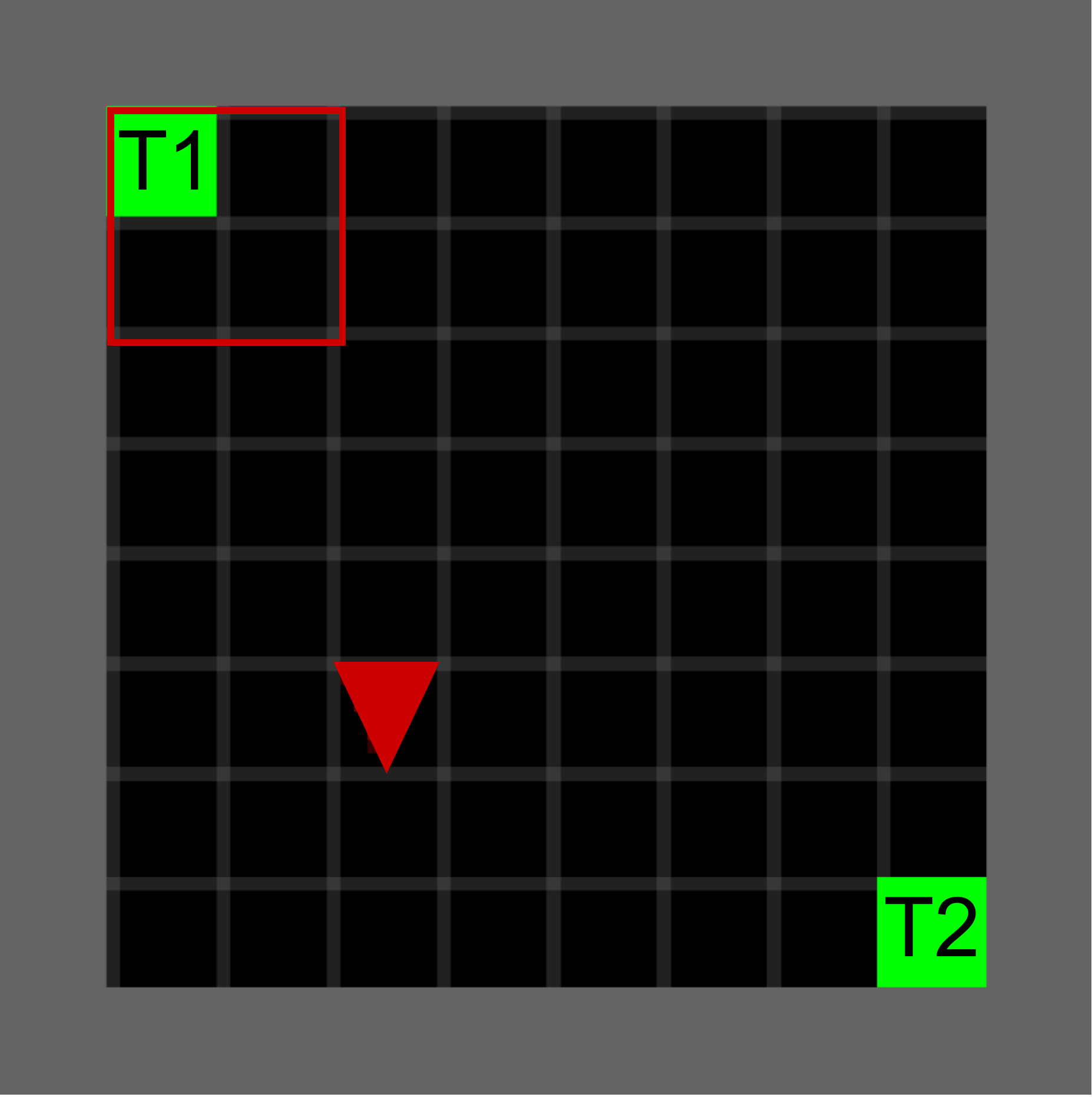}
        \caption{MiniGridLoCA} \label{fig:minigrid_illus}
    \end{subfigure}
    \caption{Illustration of the \mcloca and \mgloca setups. The solid red lines indicate the T1-zone boundaries in Phase 2 of the \loca setup.}
    \vspace{-0.25cm}
    \label{fig:nonlinear_dyna_env_illus}
\end{wrapfigure}

\textbf{Environmental Details.} We perform our evaluation on the \loca setup of two domains: (i) the MountainCar domain (MountainCarLoCA), and (ii) a simple MiniGrid domain (MiniGridLoCA) \citep{minigrid}. We choose the domains as they were also used in the experiments of previous studies regarding the \loca setup \citep{van2020loca, wan2022towards, rahimi2023replay}. The \mcloca setup (Fig.\ \ref{fig:mountaincar_illus}) is built on top of the classical MountainCar domain in which an under-powered cart has to move to certain locations by taking its position and velocity information from the environment as an input. In this setup, there are two rewarding regions that act as terminal states: (i) T1 which requires the agent to be at the top of the hill and have a velocity greater than zero, and (ii) T2 which requires the agent to be at the bottom of the hill and have a velocity close to zero. The \mgloca setup (Fig.\ \ref{fig:minigrid_illus}) is built on top of a simple MiniGrid domain in which the agent has to navigate to the green goal cells by taking a top-down image of the environment as input. In this setup, there are again two rewarding regions that act as terminal states: (i) T1 which is at the top-left corner, and (ii) T2 which is at the bottom-right corner.

\begin{figure}[]
    \centering
    \begin{subfigure}{0.45\textwidth}
        \centering
        \includegraphics[width=6.25cm]{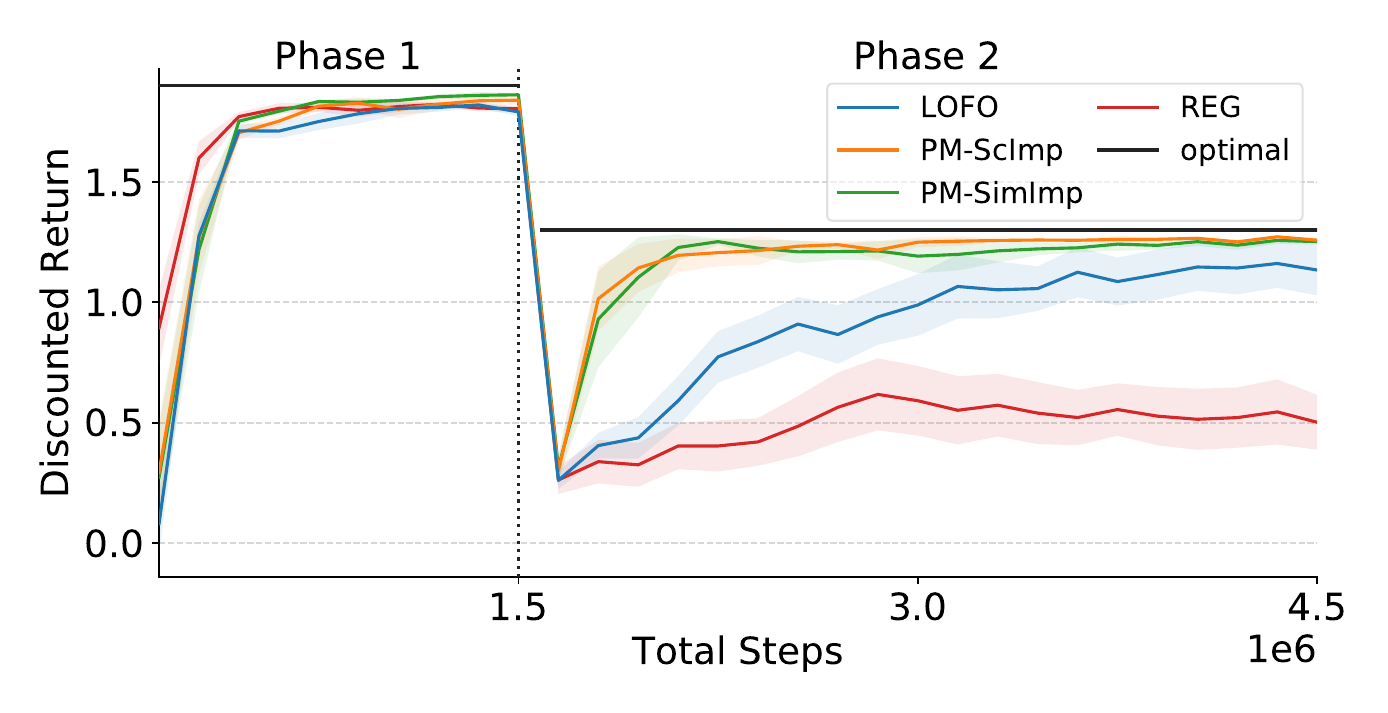}
        \caption{MountainCarLoCA} \label{fig:mountaincar_plots}
    \end{subfigure}
    \begin{subfigure}{0.45\textwidth}
        \centering
        \includegraphics[width=6.25cm]{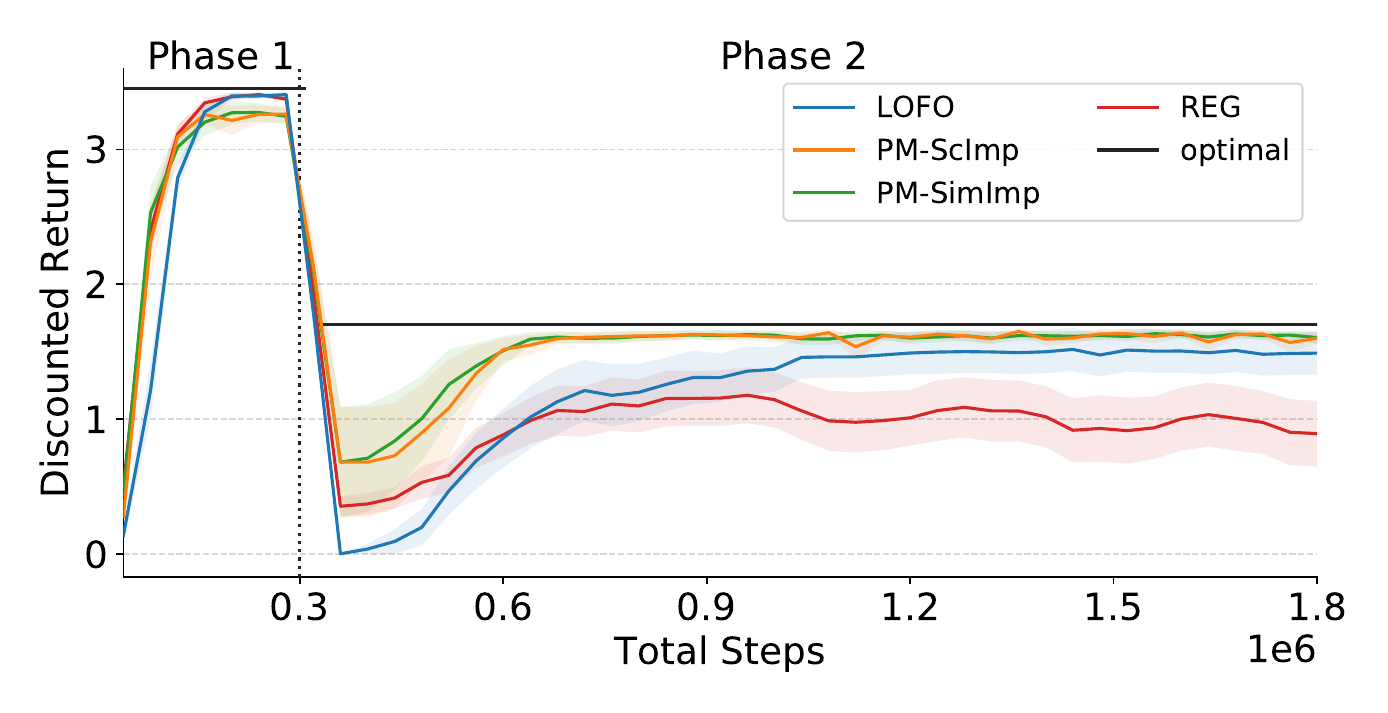}
        \caption{MiniGridLoCA} \label{fig:minigrid_plots}
    \end{subfigure}
    \caption{Plots showing the learning curves of the deep Dyna-Q agents that are referred to as PM-SimImp, PM-ScImp, REG and LOFO on the (a) \mcloca and (b) \mgloca setups. Each learning curve is an average discounted return over 20 runs and the shaded area represents the confidence intervals. The maximum possible return in each phase is represented by a solid black line.}
    \label{fig:nonlinear_dyna}
\end{figure}

\textbf{Deep Dyna-Q with Partial Models and Baselines.} In order to demonstrate the flexibility in implementing partial models, we implement two deep Dyna-Q agents with varying partial model implementations: (i) in the first one the agent employs a simple implementation of partial models (PM-SimImp), and (ii) in the second one the agent employs a scalable implementation of partial models (PM-ScImp). We compare these agents with two baseline agents from \citet{rahimi2023replay}: (i) a regular deep Dyna-Q agent (REG), and (ii) a deep Dyna-Q agent with a specifically designed replay buffer that removes the oldest sample from a local neighbourhood of a new sample in its replay buffer to address the interference-forgetting dilemma (LOFO). For both of these baselines, we choose the best performing agents from \citet{rahimi2023replay}. 

\textbf{Quantitative Analysis.} Fig.\ \ref{fig:mountaincar_plots} \& \ref{fig:minigrid_plots} show the learning curves of the different deep Dyna-Q agents on the MountainCarLoCA and MiniGridLoCA setups, respectively. As can be seen, while all the agents reach close to optimal performance in Phase 1, the REG agent fails in adapting to the local change in Phase 2 as it suffers from the interference-forgetting dilemma. And, even though the PM-SimImp, PM-ScImp and LOFO agents are all able to display adaptability in Phase 2, which makes all of them adaptive MBRL agents as per the LoCA setup, the PM-SimImp and PM-ScImp agents are able adapt much faster compared to the LOFO agent, demonstrating that besides addressing the interference-forgetting dilemma and the proper model update challenge, they also address the quick adaptation challenge. In order to demonstrate the generality of the performance of partial models across different LoCA setups, we also evaluated the different deep Dyna-Q agents on the LoCA1 and LoCA2 setups of the MountainCar and MiniGrid domains. Results in Fig. \ref{fig:nonlinear_dyna_loca1} \& \ref{fig:nonlinear_dyna_loca2} show that a similar performance trend also holds in these setups.

\begin{figure}[h!]
    \centering
    \includegraphics[height=3.cm]{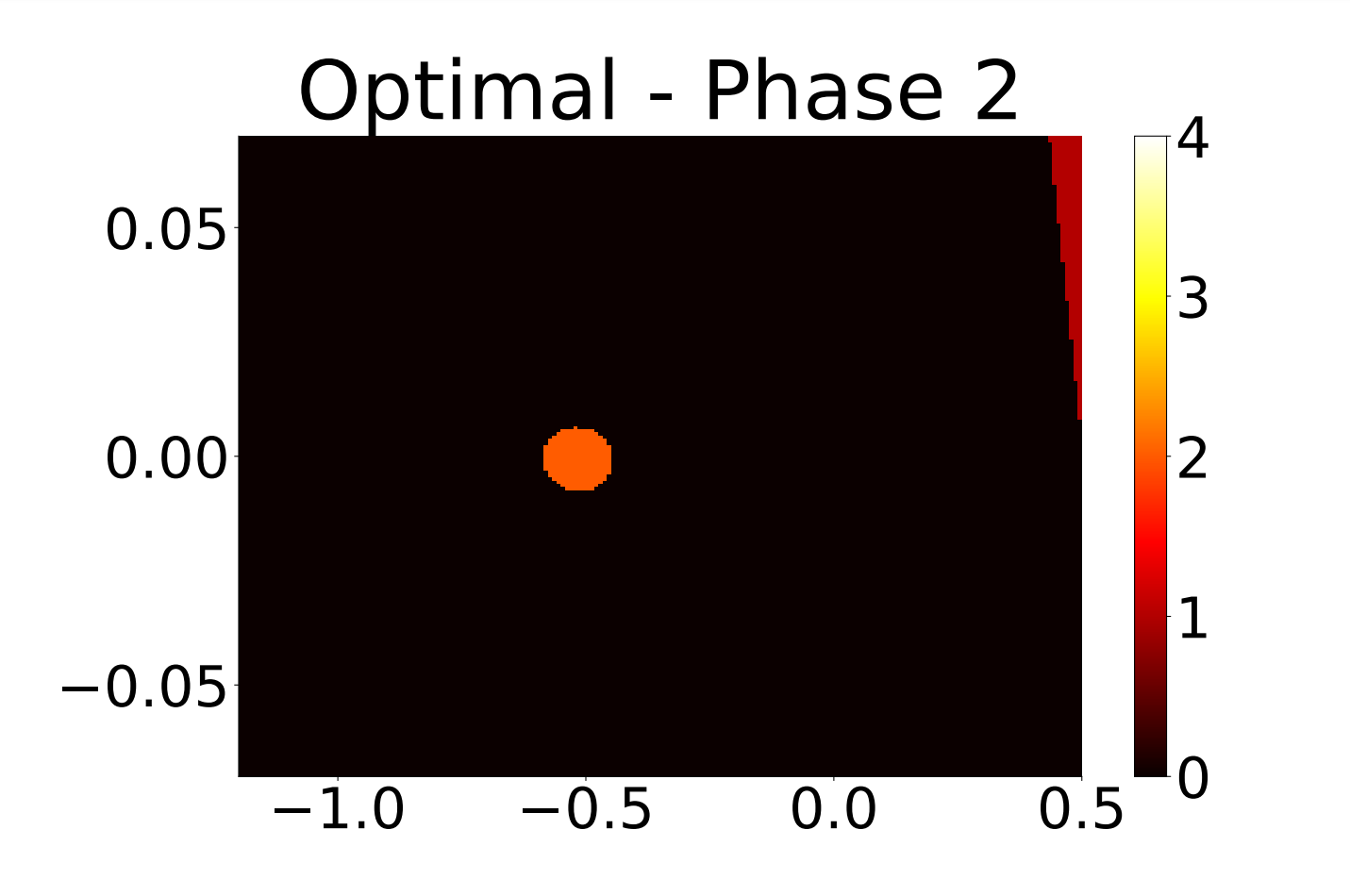}
    \includegraphics[height=3.cm]{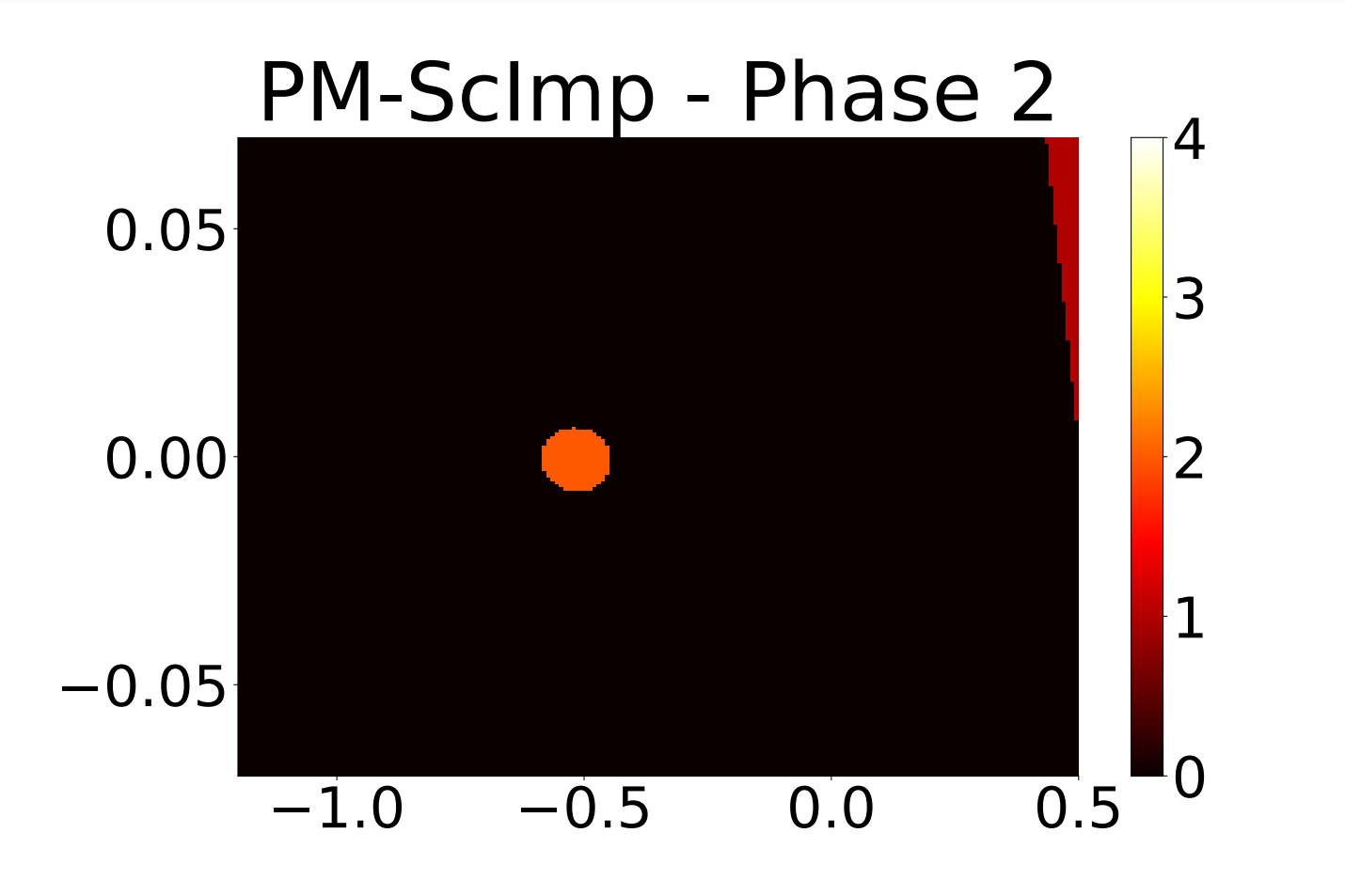}
    \includegraphics[height=3.cm]{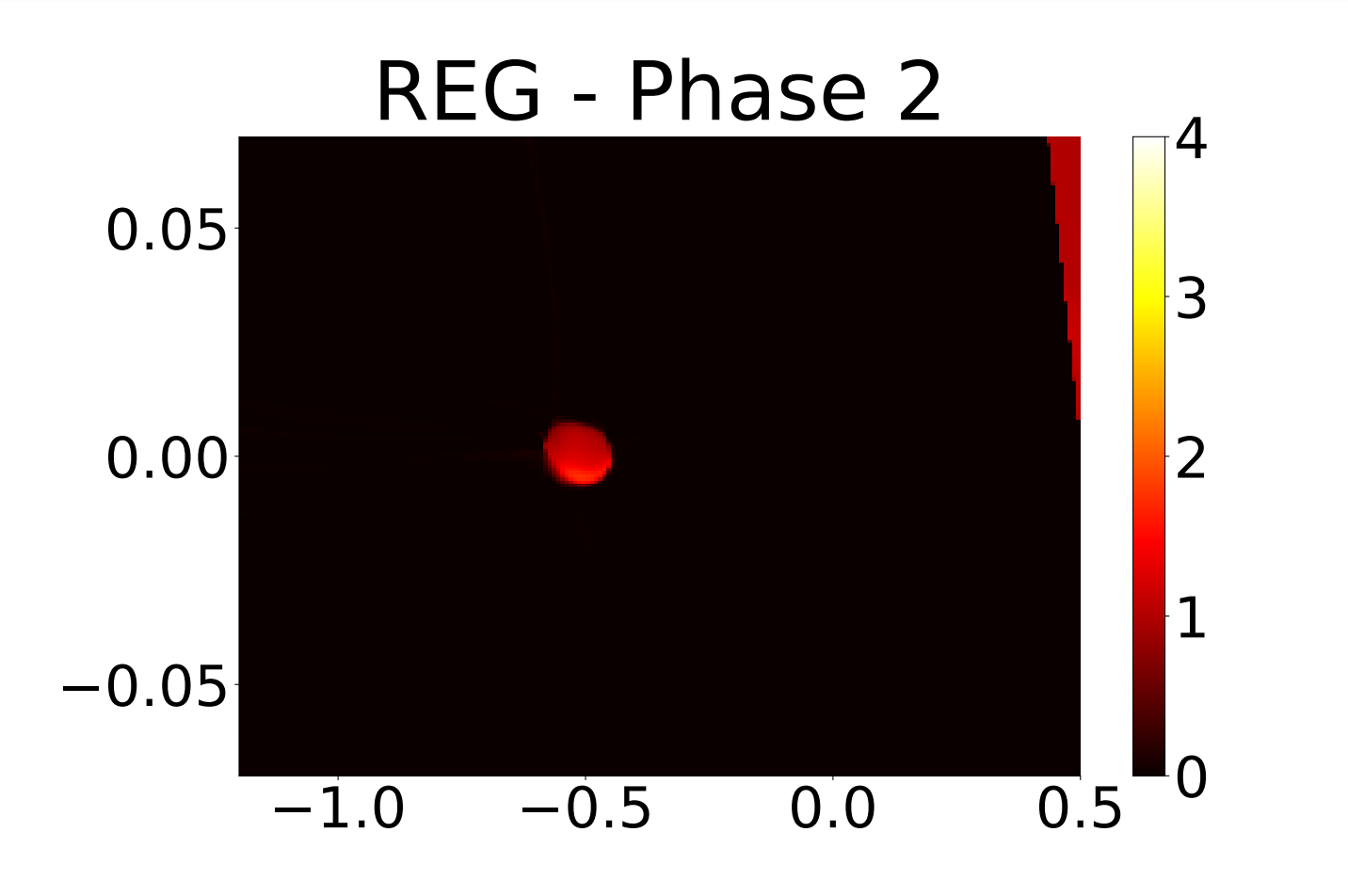}
    \caption{Visualization of the predicted rewards of the parametric models of the optimal, PM-ScImp and REG agents at the end of Phase 2 in the \mcloca setup. In each heatmap, the $x$ axis represents the agent’s position, and the $y$ axis represents its velocity.}
    \label{fig:reward_plots_mc}
\end{figure}

\textbf{Qualitative Analysis I.} To have a better understanding of how the use of partial models mitigates the interference-forgetting dilemma and the proper model update challenge, we look at the reward predictions of the PM-ScImp and REG agents at the end of Phase 2 and compare them with an optimal agent in the \mcloca setup. The visualizations are shown in Fig.\ \ref{fig:reward_plots_mc}. As can be seen, the PM-ScImp agent is able to correctly adapt its reward function to around +1 for the T1 terminal states (triangular region at the top-right corner) and maintain its reward function around +2 for the T2 terminal states (circular region in the middle), almost matching the performance of the optimal agent. On the other hand, as it is not even able to address the interference-forgetting dilemma, the REG agent fails in correctly predicting the reward function for the T2 terminal states.

\textbf{Qualitative Analysis II.} To see if the PM-SimImp and PM-ScImp agents work as intended, we also look at the number of partial models they use and the assignment places of these models in the state space. With every run, we consistently observe that these agents make use of three partial models: (i) one that is assigned to the T1 rewarding region, (ii) one that is assigned to the T2 rewarding region, and (iii) one that is assigned to the region outside of the T1 and T2 regions. For example, the reward predictions of the PM-ScImp agent (see Fig.\ \ref{fig:reward_plots_mc}) were generated by three different partial models: (i) the triangular region at the top-right corner was generated with the partial model that is responsible for the T1 terminal states, (ii) the circular region in the middle was generated with a partial model that is responsible for the T2 terminal states, and (iii) the rest was generated with a partial model that is responsible for the states outside of the T1 and T2 terminal states.

\textbf{Discussion on the Memory and Computation Costs.} In our experiments, we observed that while the experiments with PM-SimImp agent took thrice as much memory and wall clock time than the experiments with the REG agent, the experiments with PM-ScImp agent took around the same memory and wall clock time of the experiments with the REG agent. This justifies that the implementation proposed in Sec.\ \ref{sec:scalable_imp} is indeed a scalable implementation.

\section{PlaNet and Dreamer Experiments}
\label{sec:dreamer_planet}

To demonstrate the generality of the use of partial models in building adaptive deep model-based agents, we also employed partial models in modern deep MBRL agents such as \planet \citep{hafner2019learning} and \dreamer \citep{Hafner2020Dream, hafner2021mastering, hafner2023mastering} and evaluated their performance in the \loca setup. The details of all of the experiments in this section can be found in App \ref{app:reacher_appendix} \& \ref{app:randomreacher_appendix}. 

\begin{wrapfigure}{R}{0.5\textwidth}
    \centering
    \begin{subfigure}{0.23\textwidth}
        \centering
        \includegraphics[width=2.5cm]{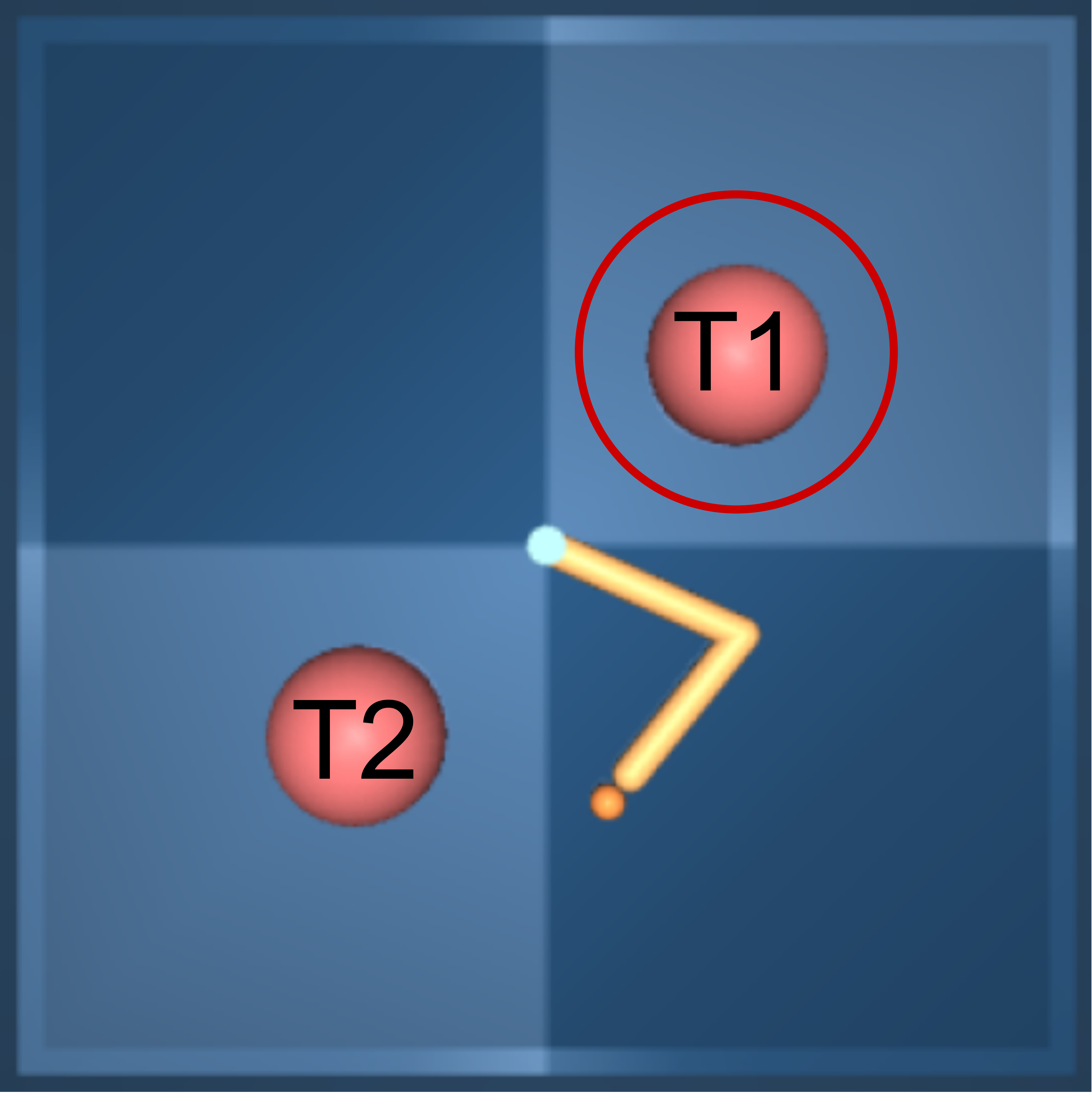}
        \caption{ReacherLoCA} \label{fig:reacher_illus}
    \end{subfigure}
    \begin{subfigure}{0.23\textwidth}
        \centering
        \includegraphics[width=2.5cm]{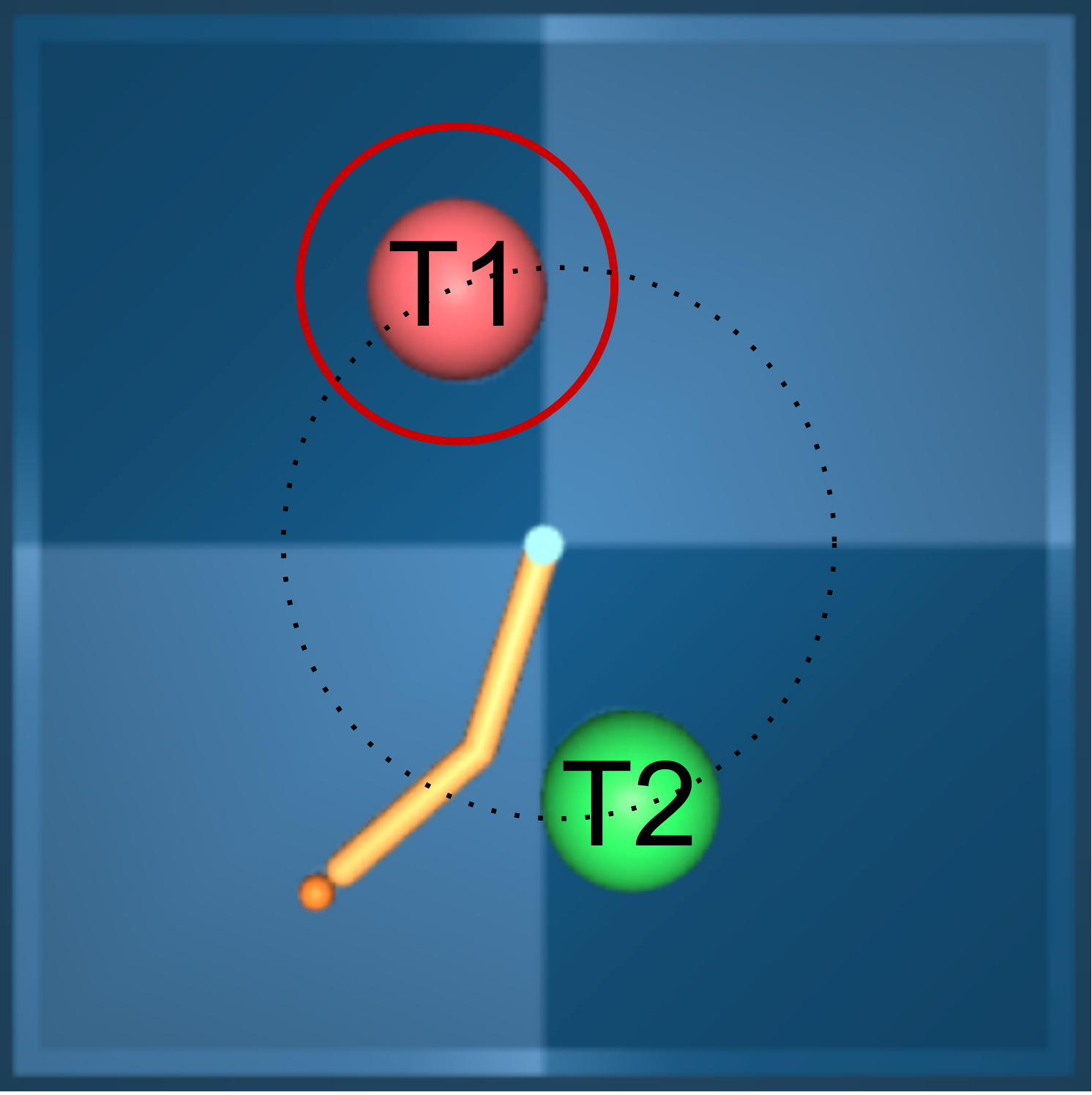}
        \caption{RandomReacherLoCA} \label{fig:rand_reacher_illus}
    \end{subfigure}
    \caption{Illustration of the \reachloca and \rndreachloca setups. The solid red lines indicate the T1-zone boundaries in Phase 2 of the \loca setup.}
    \vspace{-0.25cm}
    \label{fig:modern_env_illus}
\end{wrapfigure}

\textbf{Environmental Details.} We evaluate these agents on the \loca setup of two domains: (i) the pixel-based Reacher domain that was introduced by \citet{wan2022towards} (ReacherLoCA), and (ii) the randomized version of the pixel-based Reacher domain (RandomReacherLoCA). Again, we choose these domains as they were used in the experiments of previous studies regarding the LoCA setup \citep{wan2022towards, rahimi2023replay}. The \reachloca setup (Fig.\ \ref{fig:reacher_illus}) is built on top of the continuous-action Reacher domain \citep{tassa2018deepmind} in which the agent has to move the tip of the outer bar to the target positions and keep it there till the end of the episode (which takes 1000 time steps) by taking an 64x64 top-down image of the environment as input. In this variation, there are two rewarding regions: (i) T1 which is at the top-right quadrant, and (ii) T2 which is at the bottom-left quadrant. The \rndreachloca setup (Fig.\ \ref{fig:rand_reacher_illus}) is a more complicated extension of the \reachloca setup where the locations of the two terminal states keep randomly changing from one episode to another, while still being exactly opposite to each other. In this setup, the T2 terminal state is colored differently from the T1 one so that the agent can differentiate between them.

\begin{figure}[]
    \centering
    \begin{subfigure}{0.31\textwidth}
        \centering
        \includegraphics[width=4.5cm]{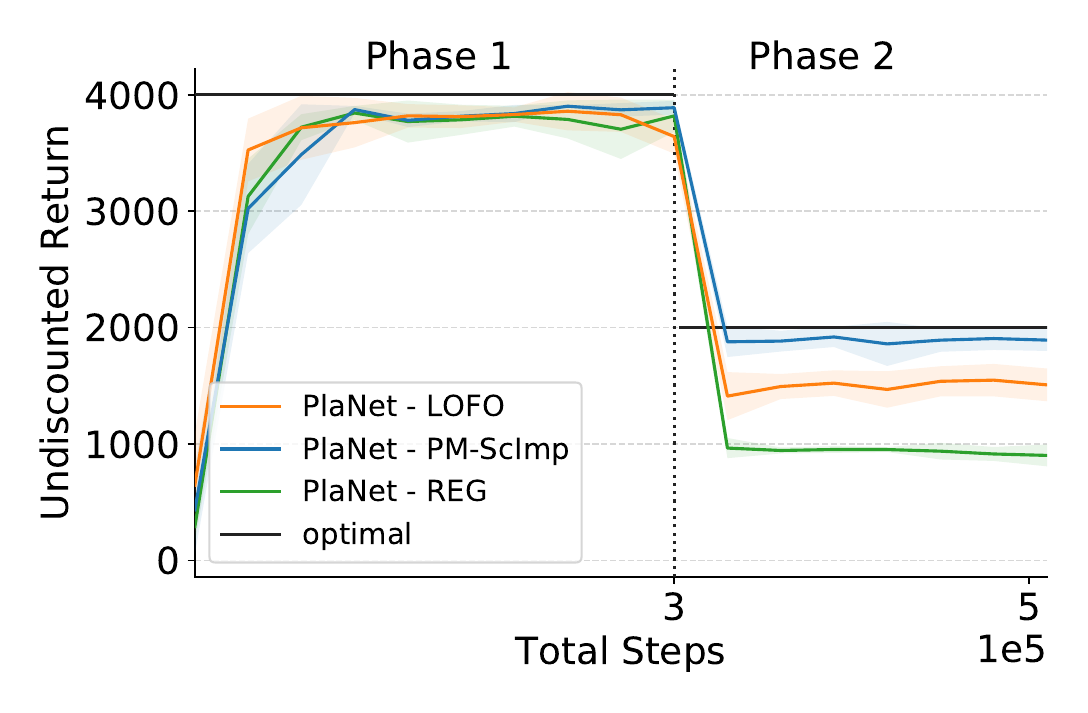}
        \caption{ReacherLoCA} \label{fig:reacherloca_planet_plots}
    \end{subfigure}
    \begin{subfigure}{0.31\textwidth}
        \centering
        \includegraphics[width=4.5cm]{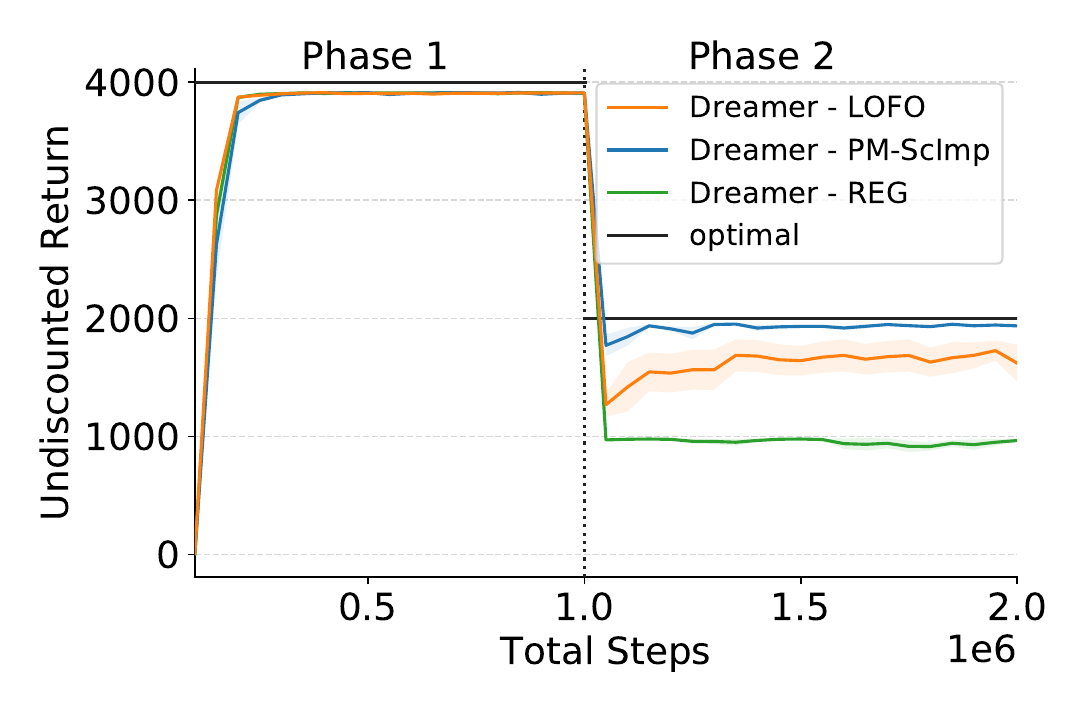}
        \caption{ReacherLoCA} \label{fig:reacherloca_dreamer_plots}
    \end{subfigure}
    \begin{subfigure}{0.31\textwidth}
        \centering
        \includegraphics[width=4.5cm]{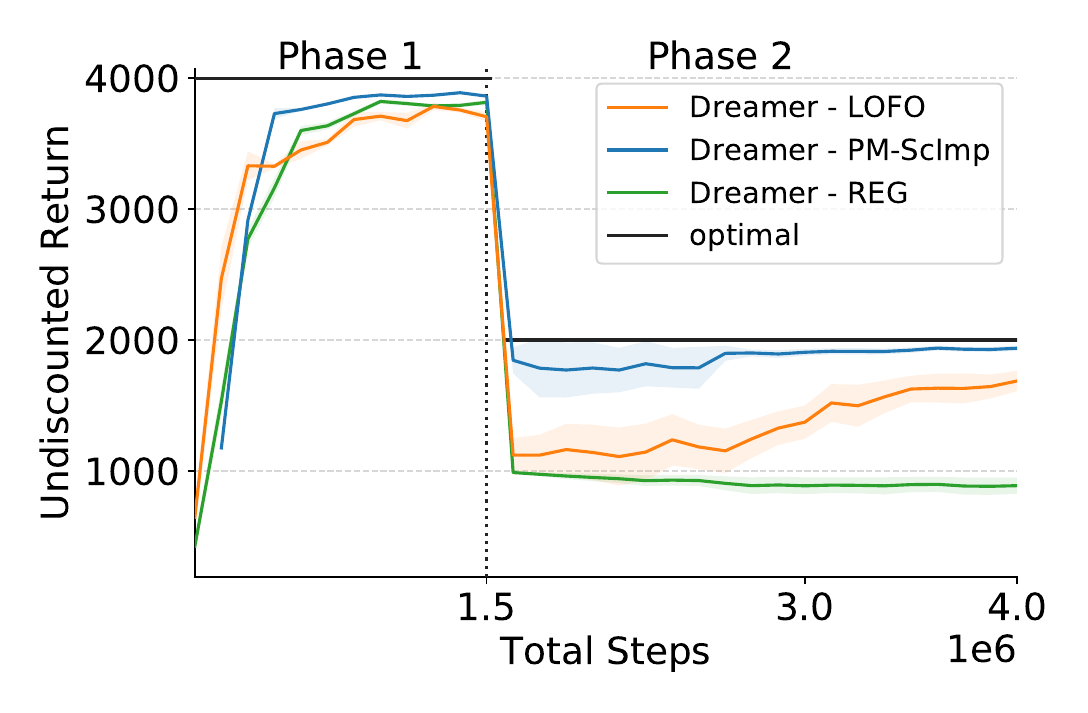}
        \caption{RandomReacherLoCA} \label{fig:random_reacherloca_dreamer_plots}
    \end{subfigure}
    \caption{Plots showing the learning curves of the PlaNet - PM-ScImp, Dreamer - PM-ScImp, PlaNet - REG, Dreamer - REG, PlaNet - LOFO, Dreamer - LOFO agents on the (a, b) \reachloca and (c) \rndreachloca setups. Each learning curve is an average undiscounted return over 10 runs and the shaded area represents the confidence intervals. The maximum possible return in each phase is represented by a solid black line.}
    \label{fig:dreamer_and_planet}
\end{figure}

\textbf{PlaNet/Dreamer with Partial Models and Baselines.} As we have already demonstrated in Sec.\ \ref{sec:nonlinear_dyna} that the two different implementations of partial models both allow for building adaptive deep MBRL agents, in this section, we only run experiments with the \planet and \dreamer agents that employ a scalable implementation of partial models (PlaNet - PM-ScImp, Dreamer - PM-ScImp). We again compare these agents with the two baseline agents from \citet{rahimi2023replay}: (i) the regular \planet and \dreamer agents (PlaNet - REG, Dreamer - REG), and (ii) the \planet and \dreamer agents with LOFO (PlaNet - LOFO, Dreamer - LOFO). Again, for both of these baselines, we choose the best performing agents from \citet{rahimi2023replay}.

\textbf{Quantitative Analysis.} Fig. \ref{fig:reacherloca_planet_plots} \& \ref{fig:reacherloca_dreamer_plots} show the learning curves of the different \planet and \dreamer agents on the \reachloca setup, and Fig. \ref{fig:random_reacherloca_dreamer_plots} shows the learning curves of the \dreamer agents on the \rndreachloca setup. We observe again that in both of the setups, while all the agents reach close to optimal performance in Phase 1, the REG agents fail in adapting to the local change in Phase 2 as they again suffer from the interference-forgetting dilemma. And, again, even though both the PM-ScImp and LOFO agents are able to display adaptability in Phase 2, which makes both of them adaptive MBRL agents as per the LoCA setup, the PM-ScImp agents are able to adapt much faster compared to the LOFO agents, again demonstrating that besides addressing the two challenges, they also address the quick adaptation challenge. In order to demonstrate the generality of the performance of partial models across different LoCA setups, we also evaluated the these agents on the LoCA1 and LoCA2 setups of the Reacher and RandomReacher domains. Results in Fig. \ref{fig:dreamer_and_planet1} \& \ref{fig:dreamer_and_planet2} show that a similar performance trend also holds in these setups.

\begin{figure}[h!]
    \centering
    \includegraphics[height=3.25cm]{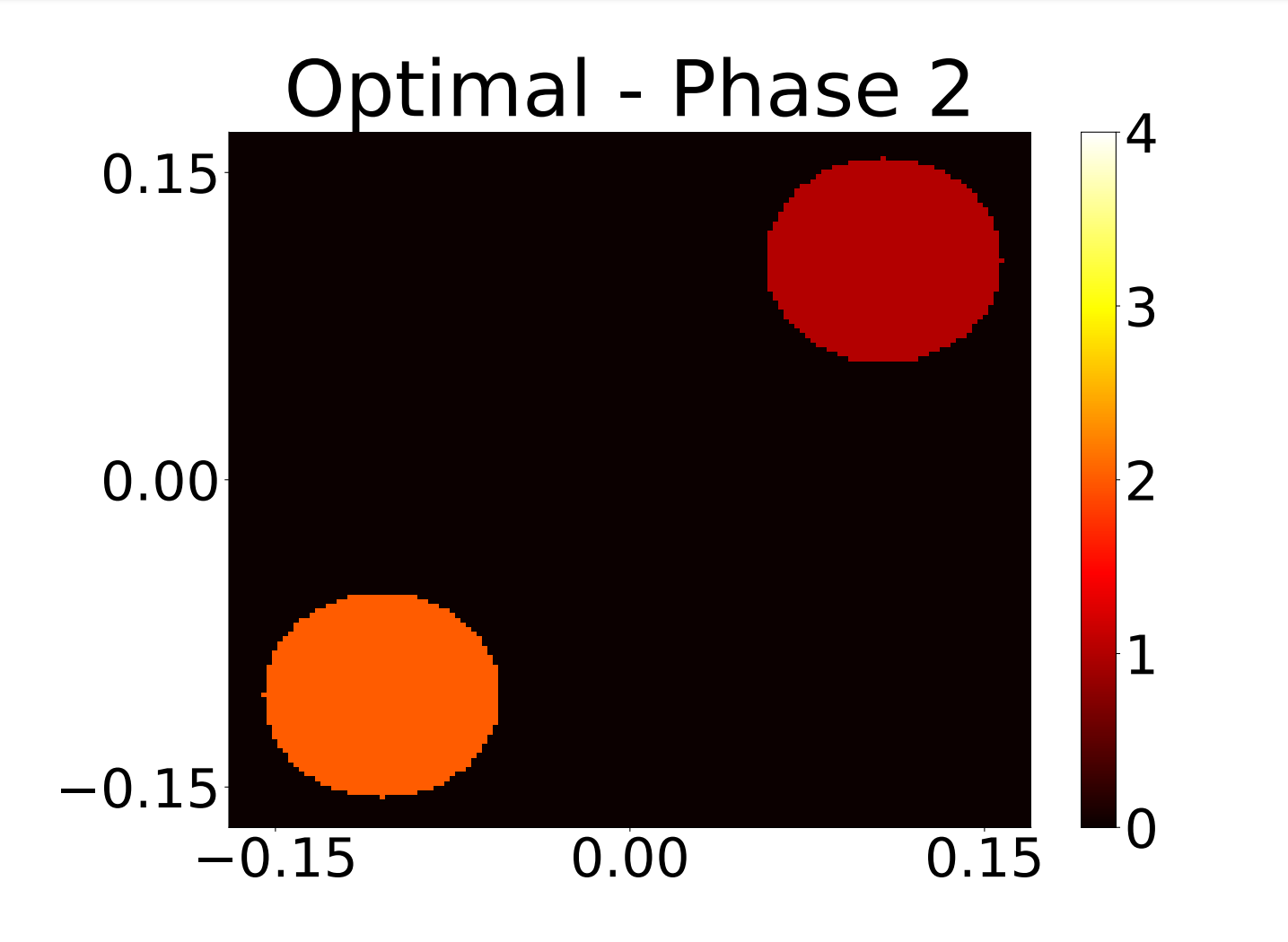}
    \includegraphics[height=3.25cm]{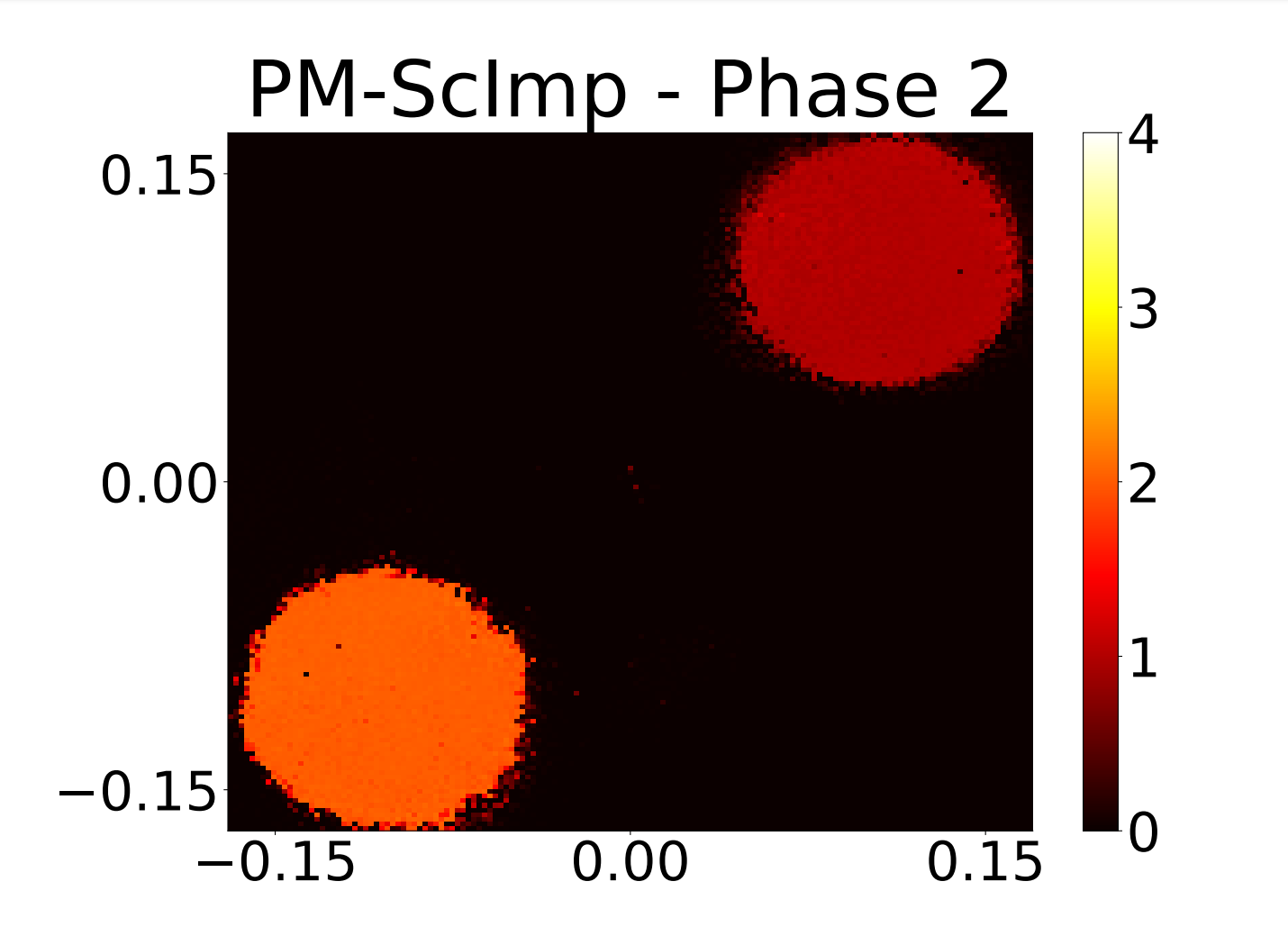}
    \includegraphics[height=3.25cm]{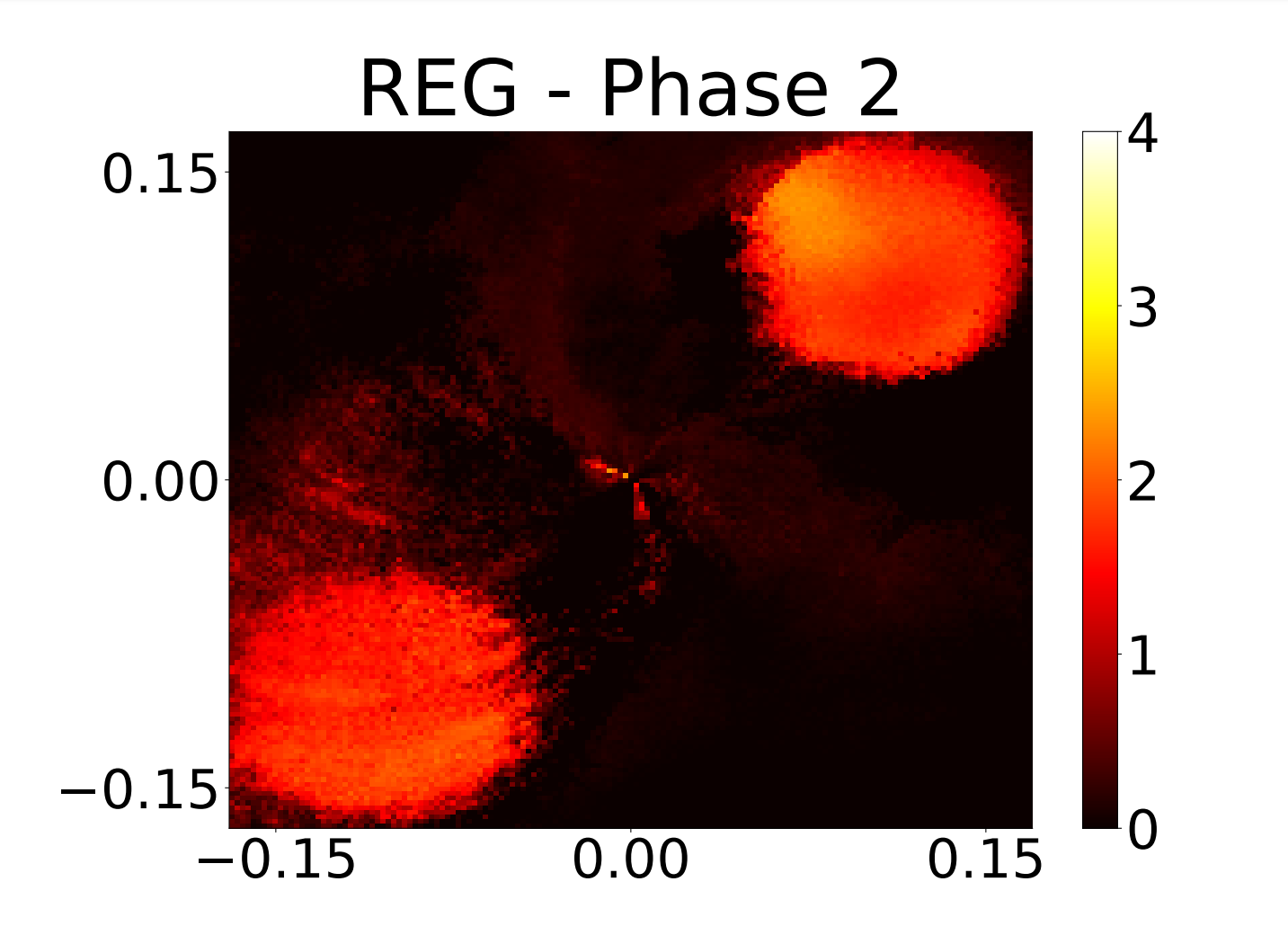}
    \caption{Visualization of the predicted rewards of the parametric models of the optimal, Dreamer - PM-ScImp and Dreamer - REG agents at the end of Phase 2 in the \reachloca setup. Each point on the heatmaps represents the agent’s position in the Reacher environment.}
    \label{fig:reward_plots_reacher}
\end{figure}

\textbf{Qualitative Analysis I.} Again, to have a better understanding of how the use of partial models mitigates the interference-forgetting dilemma and the proper model update challenge, we look at the reward predictions of the Dreamer - PM-ScImp and Dreamer - REG agents at the end of Phase 2 and compare them with an optimal agent in the \reachloca setup. The visualizations in Fig.\ \ref{fig:reward_plots_reacher} show that the PM-ScImp agent is able to correctly adapt its reward function to around +1 for the T1 rewarding region (circular region at the top-right quadrant) and maintain its reward function around +2 for the T2 rewarding regions (circular region at the bottom-left quadrant), again almost matching the optimal agent. On the other hand, as it is again not even able to address the interference-forgetting dilemma, the REG agent fails in correctly predicting the reward function for both the T1 and T2 rewarding regions.

\textbf{Qualitative Analysis II.} Again, to see if the PM-ScImp agents works as intended, we also look at the number of partial models they use and the assignment places of these models in the state space. Similar to our results in Sec.\ \ref{sec:nonlinear_dyna}, with every run, we consistently observe that these agents make use of three partial models that are assigned to the T1 rewarding region, the T2 rewarding region and the region outside of the T1 and T2 regions. For example, the reward predictions of the Dreamer - PM-ScImp agent (see Fig.\ \ref{fig:reward_plots_reacher}) were generated by three different partial models: (i) the circular region at the top-right quadrant was generated with the partial model that is responsible for the T1 terminal states, (ii) the circular region at the bottom-left quadrant was generated with a partial model that is responsible for the T2 terminal states, and (iii) the rest was generated with a partial model that is responsible for the states outside of the T1 and T2 terminal states.

\textbf{Discussion on the Memory and Computation Costs.} In our experiments, we observed again that the experiments with PM-ScImp agent took around the same memory and wall clock time of the experiments with the REG agent, which again justifies that the scalable implementation of partial models (Sec.\ \ref{sec:scalable_imp}) is indeed scalable in terms of the memory and computation requirements.

\section{Related Work}
\label{sec:related_work}

\textbf{Adaptability in Model-Based RL.} Our study focuses on building adaptive deep MBRL agents in the \loca setup, therefore it is mostly related to the studies of \citet{van2020loca} and \citet{wan2022towards}. As mentioned previously, while the former of these studies introduced the setup, the latter one improved it in several ways (see Sec.\ \ref{sec:background}). Our study is also related to the study of \citet{rahimi2023replay} which proposed the use of a special kind of replay buffer, called LOFO, for achieving adaptivity in the LoCA setup. This replay buffer overcomes the interference-forgetting dilemma by first detecting the oldest sample from a local neighbourhood of a new sample and then by removing it from the replay buffer. However, in this study, rather than focusing on a solution on the replay buffer side, we focused on a one that exploits the power of having modularity in the model structure. We also note that, unlike \citet{rahimi2023replay}, our solution does not rely on performing a search through the whole replay buffer at every time step, which brings computational complexity benefits to our approach over LOFO.

\textbf{Partial Models in RL.} In the context of RL, partial models are models that are over certain aspects of the environment. In the literature, various forms of partial models have been considered: e.g. (i) \cite{talvitie2008simple} considers models that are partial at the pixel level, i.e. the learned model only models certain pixels in the input observation, (ii) \cite{khetarpal2021temporally} considers models that are partial at the action level, i.e. the learned model only models the consequences of certain actions, and (iii) \cite{zhao2021consciousness} and  \cite{alver2023minimal} consider models that are partial at the feature level, i.e. the learned model only models the evolution of certain features. Inspired by these studies, our study considers models that are partial at the state level, i.e. the learned models only model the transitions and reward functions in certain regions of the state space. However, unlike prior studies, in our study the MBRL agent maintains multiple partial models and effectively uses them for achieving local adaptivity. It is also important to note that, in this study, the use of partial models is explicitly for the purposes of adaptivity, and not for gaining any benefits in the computational complexity of the planning process like in \cite{khetarpal2021temporally} and \cite{alver2023minimal}.

\textbf{Ensemble Learning in RL.} As our approach makes use of multiple partial models, it can be considered as an ensemble learning method. However, while there has been several studies that make use of ensemble learning methods in RL \citep{song2023ensemble}, our approach differs in that we apply ensemble learning techniques to model-based RL algorithms.

\textbf{The Continual RL Problem.} The LoCA setup measures the adaptivity of an agent and thus serves as a preliminary yet important step towards the ambitious continual RL problem \citep{parisi2019continual, khetarpal2022towards, kessler2022same}. It specifies a particular kind of continual RL problem which involves a local environmental change. However, it should also be noted that, unlike regular continual RL problems where the agent has to prevent catastrophic forgetting and perform well across all the tasks that it has seen so far, in the LoCA setup the only thing that is of importance is the agent's performance in the current task.

\textbf{The Transfer Learnıng Problem.} As there is a need to solve multiple tasks, the LoCA setup is also related to the well-known transfer learning problem \citep{lazaric2012transfer, taylor2009transfer, zhu2023transfer}. However, unlike the transfer learning problem, in the LoCA setup the agent is not informed of which task it has to solve.

\textbf{Model-Based RL for Contınual RL.} While there has been several studies that focus on developing MBRL algorithms for the continual learning and transfer learning problems \citep{zhang2019solar, huang2021continual, nguyen2012transferring, boloka2021knowledge}, none of these algorithms directly address the challenges that were depicted in Sec.\ \ref{sec:challenges}. Therefore, generally speaking, they are not likely to demonstrate adaptivity in the LoCA setup.

\section{Conclusion and Discussion}
\label{sec:conc_and_disc}

In this study, we focused on one of the key characteristics of model-based behavior: the ability to adapt to local changes in the environment. After discussing the three main challenges for performing adaptive deep MBRL, we proposed the use of partial models as a way to mitigate these challenges. We provided two specific implementations, a simple and a scalable one, and demonstrated through experiments that when employed in state-of-the-art deep model-based agents, such as PlaNet and Dreamer, these models do not only allow for adaptivity to the local changes but also allow for quick adaptation to such changes. We believe that this is an important step towards the ambitious goal of building continual RL agents as continual RL scenarios often require adaptation to changes in the environment. 

It is important to note that even though we have only considered specific local adaptation setups (see Sec.\ \ref{sec:add_loca_setups}), the idea of having \emph{modularity in the model} of the agent is a general idea that can also be leveraged in other (local or non-local) adaptation setups. Finally, we note that, similar to previous studies in the literature \citep{rahimi2023replay}, our study also assumes (i) a good initial exploration phase and (ii) a good way to measure similarity between states, which we do with their corresponding rewards, so that it is possible to learn neural embeddings of them.

\subsubsection*{Acknowledgments}
This project has been partly funded by an NSERC Discovery grant and the Canada-CIFAR AI Chair program. We would like to thank the anonymous reviewers for providing critical and constructive feedback.

\bibliography{collas2024_conference}
\bibliographystyle{collas2024_conference}

\clearpage
\appendix

\section{Pseudocode of the ICOC Algorithm}
\label{app:icoc}

The detailed pseudocode of the \textsc{ICOC} algorithm for identifying the clusters and obtaining a classifier is depicted in Alg.\ \ref{alg:icoc_alg}. Note that the details of how this pseudocode is actually implemented can change from one agent to the other (e.g. Deep Dyna-Q, Dreamer, etc.), however, the main structure will remain the same.

\begin{algorithm}[]
    \footnotesize
    \centering
    \caption{Detailed pseudocode for identifying the clusters and obtaining a classifier.} \label{alg:icoc_alg}
    \begin{algorithmic}[1]
        \Procedure{IdentifyClusters\&ObtainAClassifier}{$\mathcal{D}$}
            \State $E\gets $ learn with $\mathcal{D}$ a neural embedding function of states such that states that have a similar reward are also closer in their \Statex \hspace*{15mm}neural embedding representation (e.g.\ by using the contrastive learning method used in \citet{rahimi2023replay})
            \State $\mathcal{D}_u\gets $ form an unlabelled dataset of vectors by concatenating the neural embedding of the state $E(S)$ and the reward $R$ of \Statex \hspace*{15mm}each transition in $\mathcal{D}$
            \State $n \gets $ identify the optimal number of clusters by running a clustering algorithm over the neural embeddings part of $\mathcal{D}_u$ (e.g.\ \Statex \hspace*{15mm}by using k-means clustering and the elbow method)
            \State $\mathcal{CC} \gets $ create a dictionary where the keys are the cluster IDs and the values are the corresponding neural embedding and \Statex \hspace*{15mm}reward components of the cluster centers ($\mathcal{CC}$ comes from cluster centers)
            \State $\mathcal{D}_l \gets $ form a labeled dataset of $\mathcal{D}_u$ by using their cluster IDs (i.e. labels are the cluster IDs)
            \State $\mathcal{D}_a \gets $ create an artificial version of $\mathcal{D}_u$ in which the reward components of the vectors are replaced with artificial rewards \Statex \hspace*{15mm}that are not seen during training and label these vectors as ``anomalous''
            \State $\mathcal{D}_l \gets \mathcal{D}_l + \mathcal{D}_a$ (unify the two datasets into one)
            \State $C\gets$ train a classifier over $\mathcal{D}_l$
            \State \textbf{return} $\mathcal{CC}$, $C$, $E$, $n$
        \EndProcedure
    \end{algorithmic}
\end{algorithm}

\section{Details of the MountainCarLoCA Experiments}
\label{app:mountaincar_appendix}

\subsection{Experimental Setup}

In the MountainCarLoCA setup \citep{van2020loca}, T1 is located at the top of the mountain ($\text{position} > 0.5$, and $\text{velocity} > 0$), and T2 is located at the valley ($(\text{position} + 0.52)^2 + 100 \times \text{velocity}^2 \leq 0.07^2$). The boundary for the initial state distribution encircles all the states within $0.4 \leq \text{position} \leq 0.5$ and $0 \leq \text{velocity} \leq 0.07$. The discount factor is $\lambda = 0.99$. For each evaluation, the agent is initialized roughly in the middle of T1 and T2 ($-0.2 \leq \text{position} \leq -0.1$ and $-0.01 \leq \text{velocity} \leq 0.01$). Table \ref{tab:Experiment_Setup_Deep_DynaQ_MountainCarLoCA} shows the details of the experimental setup that was used to evaluate the adaptivity of the deep Dyna-Q agents in MountainCarLoCA.

\subsection{Hyperparameters}

In our experiments, we used the same deep Dyna-Q agent that was used by \citet{wan2022towards} (we refer to it as REG in this study). However, similar to \cite{rahimi2023replay}, instead of having separate neural networks for different actions in each part of the model (transition, reward, and termination models), we use just one network and concatenate the action with the output of the middle layer (the one that has 63 output units) and then feed it to the next layer. Table \ref{tab:Hyperparameters_Deep_DynaQ_MountainCarLoCA} summarizes the important hyperparameters of the REG agent. Note that these are the same hyperparameters that were used by \citet{rahimi2023replay}. For the PM-SimImp and PM-ScImp agents we have just extended the REG agent in a way that is described in Sec.\ \ref{sec:partial_models} and thus the hyperparameters are the same as the REG agent.

As the LOFO agent is also built on top of REG agent, the base hyperparameters of it is again the same as the REG agent and the additional hyperparameters of it can be reached at Table \ref{tab:LoFo_MountainCarLoCA_Params}. Note again that these are the same hyperparameters that were used by \citet{rahimi2023replay}. For our neural embedding function, we have used the same hyperparameters with the LOFO agent.

\begin{table}[h!]
    \footnotesize
    \centering
    \caption{Experimental setup for the deep Dyna-Q agent in MountainCarLoCA.}
    \begin{tabular}{c|c|c}
        \hline
        \multirow{7}{5em}{Initial state distributions} & Phase 1 training  & \begin{tabular}[h]{@{}c@{}}Uniform distribution over\\the entire state space\end{tabular}\\
        \cline{2-3}
        & Phase 1 evaluation  & \begin{tabular}[h]{@{}c@{}}Uniform distribution over\\a small region \end{tabular}\\
        \cline{2-3}
        & Phase 2 training  & \begin{tabular}[h]{@{}c@{}}Uniform distribution over\\ states within the red boundary\end{tabular}\\
        \cline{2-3}
        & Phase 2 evaluation  & \begin{tabular}[h]{@{}c@{}}Uniform distribution over\\a small region \end{tabular}\\
        \hline
        \multirow{2}{5em}{Training steps} & Phase 1 steps & $1.5 \times 10^6$\\
        & Phase 2 steps & $3 \times 10^6$\\
        \hline
        \multirow{4}{5em}{Other details} 
        & \begin{tabular}[h]{@{}c@{}}Maximum number of steps\\ before an episode terminates\end{tabular} & 500\\
        & Training steps between two evaluations & $10^4$\\
        & Number of runs & 5\\
        & Number of evaluation episodes & 10\\
    \hline
    \end{tabular}
    \label{tab:Experiment_Setup_Deep_DynaQ_MountainCarLoCA}
\end{table}

\begin{table}[h!]
    \footnotesize
    \centering
    \caption{Hyperparameters of the deep Dyna-Q agent in MountainCarLoCA.}
    \begin{tabular}{c|c|c}
        \hline
        \multirow{6}{4em}{Neural networks} & Transition model  & \begin{tabular}[h]{@{}c@{}}MLP with \textit{tanh},\\$[64 \times 64 \times 63 \times 64 \times 64]$, \end{tabular}\\
        \cline{2-3}
        & Reward model  & \begin{tabular}[h]{@{}c@{}}MLP with \textit{tanh},\\$[64 \times 64 \times 63 \times 64 \times 64]$, \end{tabular}\\
        \cline{2-3}
        & Termination model  & \begin{tabular}[h]{@{}c@{}}MLP with \textit{tanh},\\$[64 \times 64 \times 63 \times 64 \times 64]$, \end{tabular}\\
        \cline{2-3}
        & Action-value estimator  & \begin{tabular}[h]{@{}c@{}}MLP with \textit{tanh},\\$[64 \times 64 \times 64 \times 64]$, \end{tabular}\\
        \hline
        \multirow{3}{4em}{Optimizer} & Value optimizer & \begin{tabular}[h]{@{}c@{}}Adam,\\ learning rate: $5 \times 10^{-6}$\end{tabular}\\
        & Model optimizer & \begin{tabular}[h]{@{}c@{}}Adam,\\learning rate: $5 \times 10^{-5}$\end{tabular}\\
        \hline
        \multirow{7}{4em}{Other details} & Exploration parameter & \begin{tabular}[h]{@{}c@{}}Epsilon greedy\\ $\epsilon =0.5$\end{tabular}\\
        & \begin{tabular}[h]{@{}c@{}}Number of random steps\\ before training\end{tabular} & $50000$\\
        & \begin{tabular}[h]{@{}c@{}}Target network update frequency\end{tabular}& $500$\\
        & \begin{tabular}[h]{@{}c@{}}Number of model learning steps\end{tabular} & $5$\\
        & \begin{tabular}[h]{@{}c@{}}Number of planning steps\end{tabular} & $5$\\
        & \begin{tabular}[h]{@{}c@{}}Mini-batch size of model learning\end{tabular} & $32$\\
        & \begin{tabular}[h]{@{}c@{}}Mini-batch size of planning\end{tabular} & $32$\\
        & \begin{tabular}[h]{@{}c@{}}Replay buffer size \end{tabular} & $4.5e6$\\
    \hline
    \end{tabular}
    \label{tab:Hyperparameters_Deep_DynaQ_MountainCarLoCA}
\end{table}

\begin{table}[h!]
    \footnotesize
    \centering
    \caption{Hyperparameters of the LOFO agent in MountainCarLoCA.}
    \begin{tabular}{c|c}
        \hline
        \begin{tabular}[h]{@{}c@{}}Embedding network\\ architecture\end{tabular} &
        \begin{tabular}[h]{@{}c@{}}MLP: $[64 \times 64 \times 64 \times 16]$,\\
        Activation Function: \textit{tanh}\end{tabular}\\
        
        \begin{tabular}[h]{@{}c@{}}Optimizer\end{tabular} & Adam, learning rate: $10^{-4}$\\
        \hline
        \begin{tabular}[h]{@{}c@{}}$\beta$\end{tabular} & $10$\\
        \begin{tabular}[h]{@{}c@{}}Number of negative samples\end{tabular} & $128$\\
        \begin{tabular}[h]{@{}c@{}}Mini-batch size\end{tabular} & $32$\\
        \begin{tabular}[h]{@{}c@{}}Total number of random steps\\ for creating dataset \end{tabular} & $100000$\\
        \begin{tabular}[h]{@{}c@{}}Number of training epochs\end{tabular} & $5$\\
        \hline
        \begin{tabular}[h]{@{}c@{}}$D_{local}$\end{tabular} & $0.005$\\
        \begin{tabular}[h]{@{}c@{}}$N_{local}$\end{tabular} & $1$\\
        \hline
    \end{tabular}
    \label{tab:LoFo_MountainCarLoCA_Params}
\end{table}

\section{Details of the MiniGridLoCA Experiments}
\label{app:minigrid_appendix}

\subsection{Experimental Setup}

The MiniGridLoCA setup was implemented on top of the MiniGrid suite \citep{minigrid}. Table \ref{tab:Experiment_Setup_Deep_DynaQ_MiniGridLoCA} shows the details of the experimental setup that was used to evaluate the adaptivity of the deep Dyna-Q agents in MiniGridLoCA.

\subsection{Hyperparameters}

In our experiments, we have used similar deep Dyna-Q agents to the ones that were used in our MountainCarLoCA experiments: we only changed the neural network architecture for the model and the action-value estimator. Table \ref{tab:Hyperparameters_deep_DynaQ_Networks_MiniGridLoCA} summarizes the architecture of the neural networks of these agents. Note that, differently from the MountainCarLoCA experiments, here we first encode a given state to a low-dimensional vector for the use of other parts of the model (transition, reward, and termination model). Then, we concatenate the given action with this vector and feed it to the MLPs. Finally, Table \ref{tab:Hyperparameters_deep_DynaQ_MinigridLoCA} summarizes the important hyperparameters of the REG agent. Note that these are the same hyperparameters that were used by \citet{rahimi2023replay}. For the PM-SimImp and PM-ScImp agents we have again just extended the REG agent in a way that is described in Sec.\ \ref{sec:partial_models} and thus the hyperparameters are the same as the REG agent.

As the LOFO agent is also built on top of REG agent, the base hyperparameters of it is again the same as the REG agent and the additional hyperparameters of it can be reached at Table \ref{tab:LoFo_MiniGridLoCA_Params}. Note again that these are the same hyperparameters that were used by \citet{rahimi2023replay}. For our neural embedding function, we have used the same hyperparameters with the LOFO agent.

\begin{table}[h!]
    \footnotesize
    \centering
    \caption{Experimental setup for the deep Dyna-Q agent in MiniGridLoCA.}
    \begin{tabular}{c|c|c}
        \hline
        \multirow{8}{5em}{Initial state distributions} & Phase 1 training  & \begin{tabular}[h]{@{}c@{}}Uniform distribution over\\the entire state space\end{tabular}\\
        \cline{2-3}
        & Phase 1 evaluation  & \begin{tabular}[h]{@{}c@{}}Uniform distribution over\\ the entire state space\end{tabular}\\
        \cline{2-3}
        & Phase 2 training  & \begin{tabular}[h]{@{}c@{}}Uniform distribution over\\ states within the red boundary \\($2 \times 2$ subgrid) \end{tabular}\\
        \cline{2-3}
        & Phase 2 evaluation  & \begin{tabular}[h]{@{}c@{}}Uniform distribution over\\ the entire state space\end{tabular}\\
        \hline
        \multirow{2}{5em}{Training steps} & Phase 1 steps & $3 \times 10^5$\\
        & Phase 2 steps & $1.5 \times 10^6$\\
        \hline
        \multirow{4}{5em}{Other details} 
        & \begin{tabular}[h]{@{}c@{}}Maximum number of steps\\ before an episode terminates\end{tabular} & $100$\\
        & Training steps between two evaluations & $10^4$\\
        & Number of runs & $5$\\
        & Number of evaluation episodes & $10$\\
    \hline
    \end{tabular}
    \label{tab:Experiment_Setup_Deep_DynaQ_MiniGridLoCA}
\end{table}

\begin{table}[h!]
    \footnotesize
    \centering
    \caption{Neural network architecture for the deep Dyna-Q agent in MiniGridLoCA.}
    \begin{tabular}{c|c|c}
        \hline
        \multirow{20}{5em}{Neural networks} & Transition model  & \begin{tabular}[h]{@{}c@{}}CNN:\\ (Channels: $[32 \times 64 \times 64]$\\
        Kernel Sizes: $[8 \times 3 \times 3]$\\
        Strides: $[4 \times 2 \times 2]$),\\
        Followed by Transposed CNN:\\(Channels: $[64 \times 32 \times 3]$\\
        Kernel Sizes: $[6 \times 6 \times 5]$\\
        Strides: $[1 \times 4 \times 3]$),\\ Activation Function: \textit{relu} \end{tabular}\\
        \cline{2-3}
        & Reward model  & \begin{tabular}[h]{@{}c@{}}CNN:\\ (Channels: $[32 \times 64 \times 64]$\\
        Kernel Sizes: $[8 \times 3 \times 3]$\\
        Strides: $[4 \times 2 \times 2]$),\\
        Followed by MLP: $[512]$,\\ Activation Function: \textit{relu} \end{tabular}\\
        \cline{2-3}
        & Termination model  & \begin{tabular}[h]{@{}c@{}}CNN:\\ (Channels: $[32 \times 64 \times 64]$\\
        Kernel Sizes: $[8 \times 3 \times 3]$\\
        Strides: $[4 \times 2 \times 2]$),\\
        Followed by MLP: $[512]$,\\ Activation Function: \textit{relu} \end{tabular}\\
        \cline{2-3}
        & Action-value estimator  & \begin{tabular}[h]{@{}c@{}}CNN:\\ (Channels: $[32 \times 64 \times 64]$\\
        Kernel Sizes: $[8 \times 3 \times 3]$\\
        Strides: $[4 \times 2 \times 2]$),\\
        Followed by MLP: $[512]$,\\ Activation Function: \textit{relu} \end{tabular}\\
        \hline
    \end{tabular}
    \label{tab:Hyperparameters_deep_DynaQ_Networks_MiniGridLoCA}
\end{table}

\begin{table}[h!]
    \footnotesize
    \centering
    \caption{Hyperparameters of the deep Dyna-Q agent in MiniGridLoCA.}
    \begin{tabular}{c|c|c}
        \hline
        \multirow{3}{4em}{Optimizer} & Value optimizer & \begin{tabular}[h]{@{}c@{}}Adam,\\ learning rate: $6.25 \times 10^{-5}$\end{tabular}\\
        & Model optimizer & \begin{tabular}[h]{@{}c@{}}Adam,\\learning rate: $ 10^{-4}$\end{tabular}\\
        \hline
        \multirow{7}{6em}{Other details} & Exploration parameter & \begin{tabular}[h]{@{}c@{}}Epsilon greedy\\ $\epsilon = 0.5$\end{tabular}\\
        & \begin{tabular}[h]{@{}c@{}}Number of random steps\\ before training\end{tabular} & $2000$\\
        & \begin{tabular}[h]{@{}c@{}}Target network update frequency\end{tabular}& $5000$\\
        & \begin{tabular}[h]{@{}c@{}}Number of model learning steps\end{tabular} & $1$\\
        & \begin{tabular}[h]{@{}c@{}}Number of planning steps\end{tabular} & $1$\\
        & \begin{tabular}[h]{@{}c@{}}Mini-batch size of model learning\end{tabular} & $128$\\
        & \begin{tabular}[h]{@{}c@{}}Mini-batch size of planning\end{tabular} & $128$\\
        & \begin{tabular}[h]{@{}c@{}}Replay buffer size \end{tabular} & $1.8e6$\\
    \hline
    \end{tabular}
    \label{tab:Hyperparameters_deep_DynaQ_MinigridLoCA}
\end{table}

\begin{table}[h!]
    \footnotesize
    \centering
    \caption{Hyperparameters of the LOFO agent in MiniGridLoCA.}
    \begin{tabular}{c|c}
        \hline
        \begin{tabular}[h]{@{}c@{}}Embedding network\\ architecture\end{tabular} &
        \begin{tabular}[h]{@{}c@{}}CNN:\\ (Channels:$[32 \times 64 \times 64]$\\
        Kernel Sizes:$[8 \times 3 \times 3]$\\
        Strides:$[4 \times 2 \times 2]$),\\
        Followed by MLP:$[512 \times 16]$,\\ Activation Function: \textit{relu}\end{tabular}\\
        \begin{tabular}[h]{@{}c@{}}Optimizer\end{tabular} & Adam, learning rate: $10^{-4}$\\
        \hline
        \begin{tabular}[h]{@{}c@{}}$\beta$\end{tabular} & $10$\\
        \begin{tabular}[h]{@{}c@{}}Number of negative samples\end{tabular} & $128$\\
        \begin{tabular}[h]{@{}c@{}}Mini-batch size\end{tabular} & $32$\\
        \begin{tabular}[h]{@{}c@{}}Total number of random steps\\ for creating dataset \end{tabular} & $25000$\\
        \begin{tabular}[h]{@{}c@{}}Number of training epochs\end{tabular} & $5$\\
        \hline
        \begin{tabular}[h]{@{}c@{}}$D_{local}$\end{tabular} & $0.001$\\
        \begin{tabular}[h]{@{}c@{}}$N_{local}$\end{tabular} & $1$\\
        \hline
    \end{tabular}
    \label{tab:LoFo_MiniGridLoCA_Params}
\end{table}

\section{Details of the ReacherLoCA Experiments}
\label{app:reacher_appendix}

\subsection{Experimental Setup and Hyperparameters}

Tables \ref{tab:Experiment_Setup_PlaNet_MiniGridLoCA} and \ref{tab:Experiment_Setup_DreamerV2_ReacherLoCA} show the experimental setups that we used to evaluate the PlaNet and the Dreamer agents' adaptivity in ReacherLoCA, respectively.

For the hyperparameters of the PlaNet - REG and Dreamer - REG agents, we have used the same hyperparameter as in \cite{rahimi2023replay} which consist of a replay buffer size that is equal to the total amount of training steps. For the PlaNet - PM-SimImp, Dreamer - PM-SimImp, PlaNet - PM-ScImp and Dreamer - PM-ScImp agents we have just extended the corresponding REG agents in a way that is described in Sec.\ \ref{sec:partial_models} and thus the hyperparameters are the same as the corresponding REG agents.

As the PlaNet - LOFO and Dreamer - LOFO agents are also built on top of their corresponding REG agents, the base hyperparameters of them is again the same as the REG agents and the additional hyperparameters of them can be reached at Table \ref{tab:LoFo_ReacherLoCA_Params}. Note again that these are the same hyperparameters that were used by \citet{rahimi2023replay}. For our neural embedding function, we have used the same hyperparameters with the LOFO agent.

\begin{table}[h!]
    \footnotesize
    \centering
    \caption{Experimental setup for the PlaNet agent in ReacherLoCA.}
    \begin{tabular}{c|c|c}
        \hline
        \multirow{8}{5em}{Initial state distributions} & Phase 1 training  & \begin{tabular}[h]{@{}c@{}}Uniform distribution over\\the entire state space\end{tabular}\\
        \cline{2-3}
        & Phase 1 evaluation  & \begin{tabular}[h]{@{}c@{}}Uniform distribution over\\ the entire state space\end{tabular}\\
        \cline{2-3}
        & Phase 2 training  & \begin{tabular}[h]{@{}c@{}}Uniform distribution over\\ states within the red boundary\end{tabular}\\
        \cline{2-3}
        & Phase 2 evaluation  & \begin{tabular}[h]{@{}c@{}}Uniform distribution over\\ the entire state space\end{tabular}\\
        \hline
        \multirow{2}{5em}{Training steps} & Phase 1 steps & $3 \times 10^5$\\
        & Phase 2 steps & $2 \times 10^5$\\
        \hline
        \multirow{4}{5em}{Other details} 
        & \begin{tabular}[h]{@{}c@{}}Number of steps\\ before an episode terminates\end{tabular} & $1000$\\
        & Training steps between two evaluations & $15000$\\
        & Number of runs & $5$\\
        & Number of evaluation episodes & $5$\\
    \hline
    \end{tabular}
    \label{tab:Experiment_Setup_PlaNet_MiniGridLoCA}
\end{table}

\begin{table}[h!]
    \footnotesize
    \centering
    \caption{Experimental setup for the Dreamer agent in ReacherLoCA.}
    \begin{tabular}{c|c|c}
        \hline
        \multirow{8}{5em}{Initial state distributions} & Phase 1 training  & \begin{tabular}[h]{@{}c@{}}Uniform distribution over\\the entire state space\end{tabular}\\
        \cline{2-3}
        & Phase 1 evaluation  & \begin{tabular}[h]{@{}c@{}}Uniform distribution over\\ the entire state space\end{tabular}\\
        \cline{2-3}
        & Phase 2 training  & \begin{tabular}[h]{@{}c@{}}Uniform distribution over\\ states within the red boundary\end{tabular}\\
        \cline{2-3}
        & Phase 2 evaluation  & \begin{tabular}[h]{@{}c@{}}Uniform distribution over\\ the entire state space\end{tabular}\\
        \hline
        \multirow{2}{5em}{Training steps} & Phase 1 steps & $10^6$\\
        & Phase 2 steps & $10^6$\\
        \hline
        \multirow{4}{5em}{Other details} 
        & \begin{tabular}[h]{@{}c@{}}Number of steps\\ before an episode terminates\end{tabular} & $1000$\\
        & Training steps between two evaluations & $10000$\\
        & Number of runs & $5$\\
        & Number of evaluation episodes & $8$\\
    \hline
    \end{tabular}
    \label{tab:Experiment_Setup_DreamerV2_ReacherLoCA}
\end{table}

\begin{table}[h!]
    \footnotesize
    \centering
    \caption{Hyperparameters of the LOFO agent in ReacherLoCA.}
    \begin{tabular}{c|c}
        \hline
        \begin{tabular}[h]{@{}c@{}}Embedding network\\ architecture\end{tabular} &
        \begin{tabular}[h]{@{}c@{}}CNN:\\ (Channels: $[32 \times 64 \times 128 \times 256]$\\
        Kernel Sizes: $[4 \times 4 \times 4 \times 4]$\\
        Strides: $[2 \times 2 \times 2 \times 2]$),\\
        Followed by MLP: $[512 \times 64, 32]$,\\ Activation Function: \textit{relu}\end{tabular}\\
        \begin{tabular}[h]{@{}c@{}}Optimizer\end{tabular} & Adam, learning rate: $10^{-4}$\\
        \hline
        \begin{tabular}[h]{@{}c@{}}$\beta$\end{tabular} & $50$\\
        \begin{tabular}[h]{@{}c@{}}Number of negative samples\end{tabular} & $128$\\
        \begin{tabular}[h]{@{}c@{}}Mini-batch size\end{tabular} & $32$\\
        \begin{tabular}[h]{@{}c@{}}Total number of random steps\\ for creating dataset \end{tabular} & $25000$\\
        \begin{tabular}[h]{@{}c@{}}Number of training epochs\end{tabular} & $5$\\
        \hline
        \begin{tabular}[h]{@{}c@{}}$D_{local}$\end{tabular} & $0.05$\\
        \begin{tabular}[h]{@{}c@{}}$N_{local}$\end{tabular} & $10$\\
        \hline
    \end{tabular}
    \label{tab:LoFo_ReacherLoCA_Params}
\end{table}

\section{Details of the RandomReacherLoCA Experiments}
\label{app:randomreacher_appendix}

\subsection{Experimental Setup and Hyperparameters}

In the RandomReacherLoCA setup, the location of T1 (red) is randomly sampled from a circle that is centered in the middle of the environment (the dotted black circle in Fig.\ \ref{fig:rand_reacher_illus} of the main paper). And then, T2 (green) is placed at the opposite end of this circle. Table \ref{tab:Experiment_Setup_DreamerV2_RandomReacherLoCA} shows the experimental setup that we used to evaluate the Dreamer agents' adaptivity in RandomReacherLoCA.

For the hyperparameters of the PlaNet - REG and Dreamer - REG agents, we have again used the same hyperparameter as in \cite{rahimi2023replay} which consist of a replay buffer size that is equal to the total amount of training steps. For the PlaNet - PM-SimImp, Dreamer - PM-SimImp, PlaNet - PM-ScImp and Dreamer - PM-ScImp agents we have again just extended the corresponding REG agents in a way that is described in Sec.\ \ref{sec:partial_models} and thus the hyperparameters are the same as the corresponding REG agents.

As the PlaNet - LOFO and Dreamer - LOFO agents are also built on top of their corresponding REG agents, the base hyperparameters of them are again the same as the REG agents and the additional hyperparameters of them can be reached at Table \ref{tab:LoFo_ReacherLoCA_Params}. Note again that these are the same hyperparameters that were used by \citet{rahimi2023replay}. For our neural embedding function, we have used the same hyperparameters with the LOFO agent.

\begin{table}[h!]
    \footnotesize
    \centering
    \caption{Experimental setup the Dreamer agent in RandomReacherLoCA.}
    \begin{tabular}{c|c|c}
        \hline
        \multirow{8}{5em}{Initial state distributions} & Phase 1 training  & \begin{tabular}[h]{@{}c@{}}Uniform distribution over\\the entire state space\end{tabular}\\
        \cline{2-3}
        & Phase 1 evaluation  & \begin{tabular}[h]{@{}c@{}}Uniform distribution over\\ the entire state space\end{tabular}\\
        \cline{2-3}
        & Phase 2 training  & \begin{tabular}[h]{@{}c@{}}Uniform distribution over\\ states within the red boundary\end{tabular}\\
        \cline{2-3}
        & Phase 2 evaluation  & \begin{tabular}[h]{@{}c@{}}Uniform distribution over\\ the entire state space\end{tabular}\\
        \hline
        \multirow{2}{5em}{Training steps} & Phase 1 steps & $1.5 \times 10^6$\\
        & Phase 2 steps & $3.5 \times 10^6$\\
        \hline
        \multirow{4}{5em}{Other details} 
        & \begin{tabular}[h]{@{}c@{}}Number of steps\\ before an episode terminates\end{tabular} & $1000$\\
        & Training steps between two evaluations & $10000$\\
        & Number of runs & $5$\\
        & Number of evaluation episodes & $8$\\
    \hline
    \end{tabular}
    \label{tab:Experiment_Setup_DreamerV2_RandomReacherLoCA}
\end{table}

\section{Experiments on More Challenging LoCA Setups}
\label{app:challenge_loca_appendix}

\subsection{Deep Dyna-Q Experiments}
\label{app:add_loca_nonlinear_dyna_appendix}

\begin{figure}[h!]
    \centering
    \begin{subfigure}{0.45\textwidth}
        \centering
        \includegraphics[width=6.25cm]{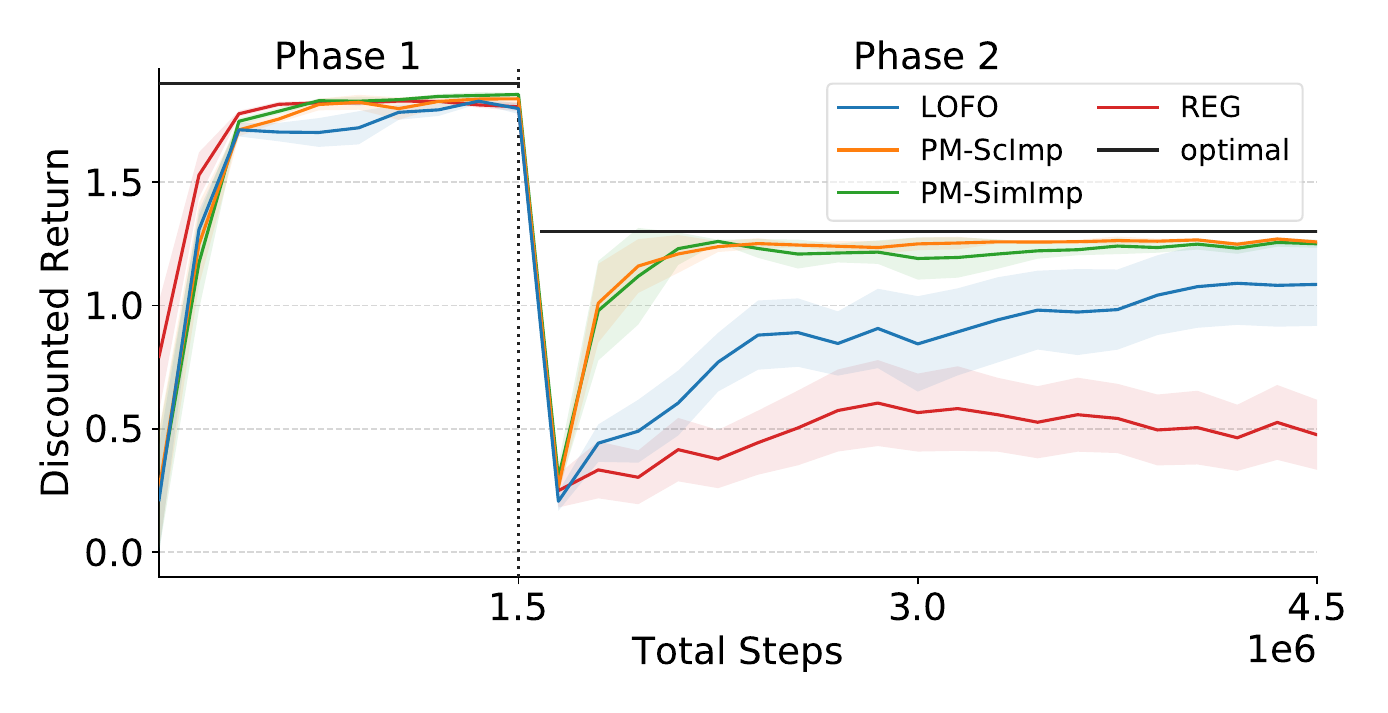}
        \caption{MountainCarLoCA} \label{fig:mountaincar_plots_loca1}
    \end{subfigure}
    \begin{subfigure}{0.45\textwidth}
        \centering
        \includegraphics[width=6.25cm]{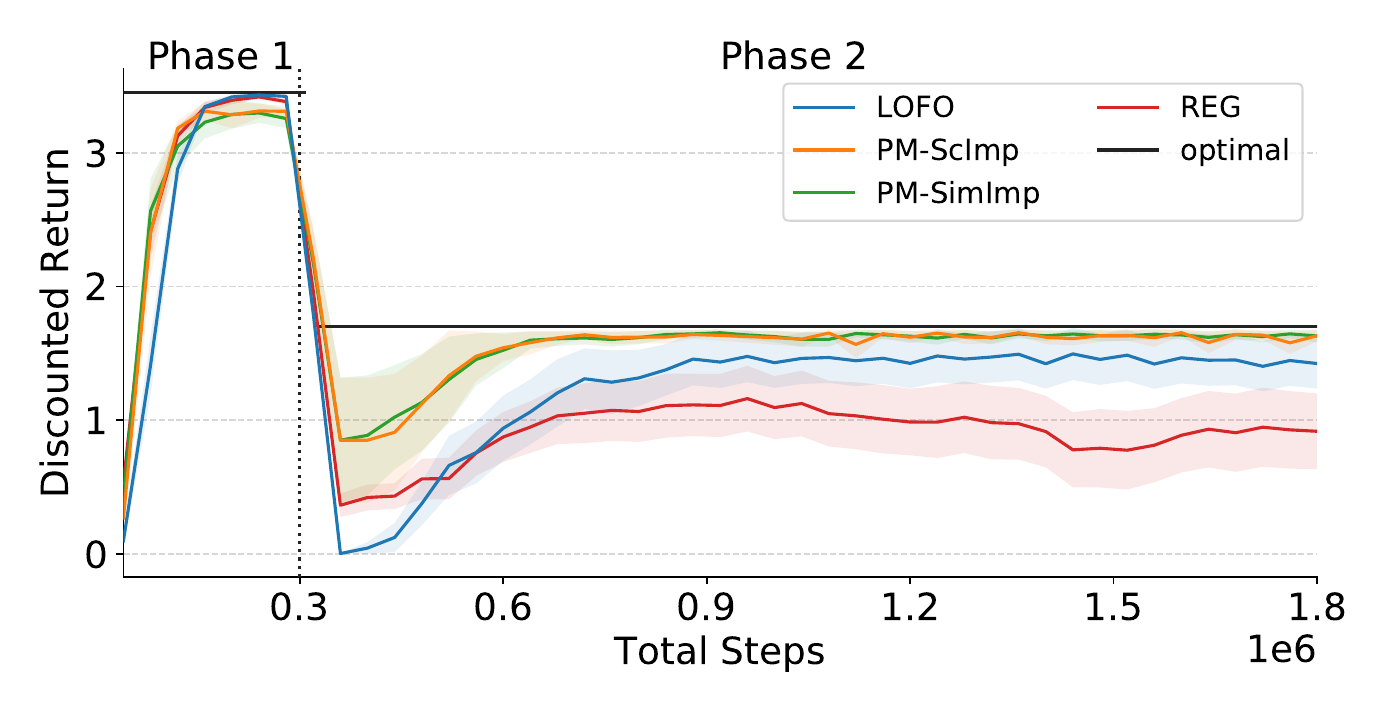}
        \caption{MiniGridLoCA} \label{fig:minigrid_plots_loca1}
    \end{subfigure}
    \caption{Plots showing the learning curves of the deep Dyna-Q agents that are referred to as PM-SimImp, PM-ScImp, REG and LOFO on the (a) MountainCarLoCA1 s and (b) MiniGridLoCA1 setups. Each learning curve is an average discounted return over 20 runs and the shaded area represents the confidence intervals. The maximum possible return in each phase is represented by a solid black line.}
    \label{fig:nonlinear_dyna_loca1}
\end{figure}

\begin{figure}[h!]
    \centering
    \begin{subfigure}{0.45\textwidth}
        \centering
        \includegraphics[width=6.25cm]{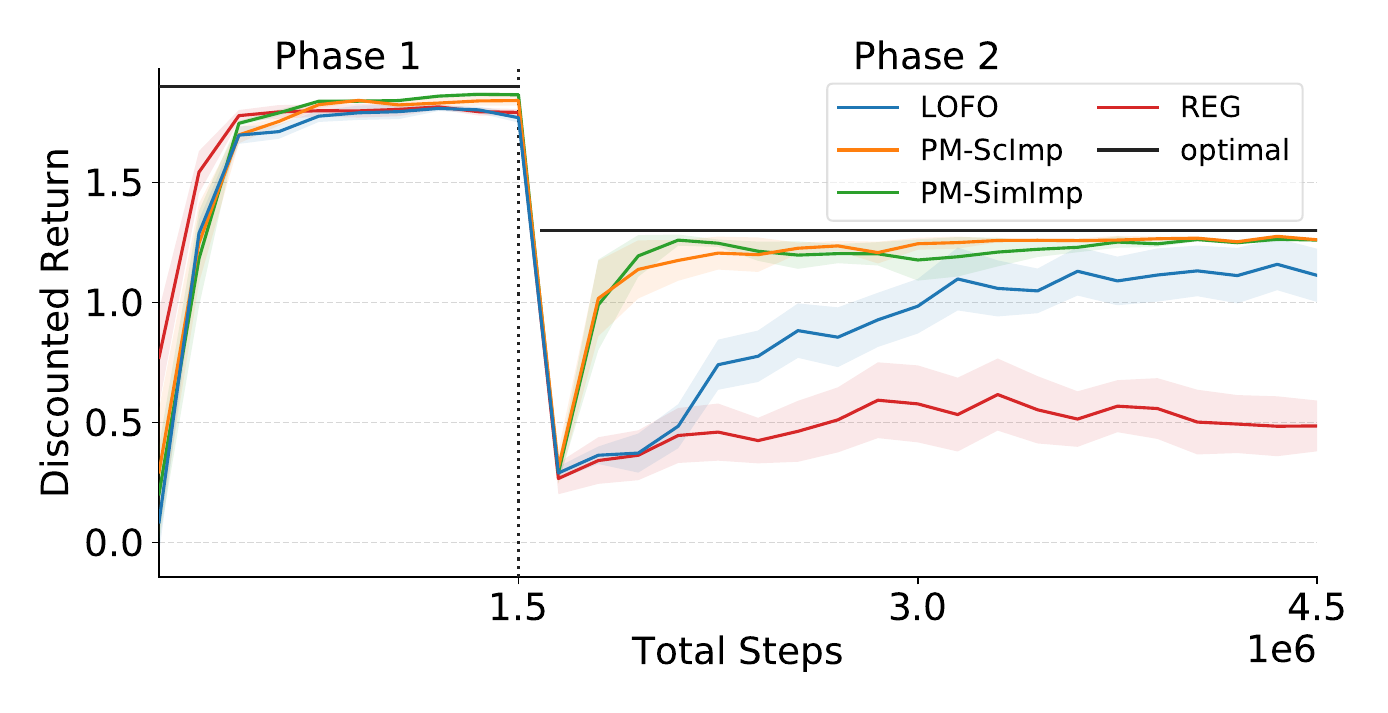}
        \caption{MountainCarLoCA} \label{fig:mountaincar_plots_loca2}
    \end{subfigure}
    \begin{subfigure}{0.45\textwidth}
        \centering
        \includegraphics[width=6.25cm]{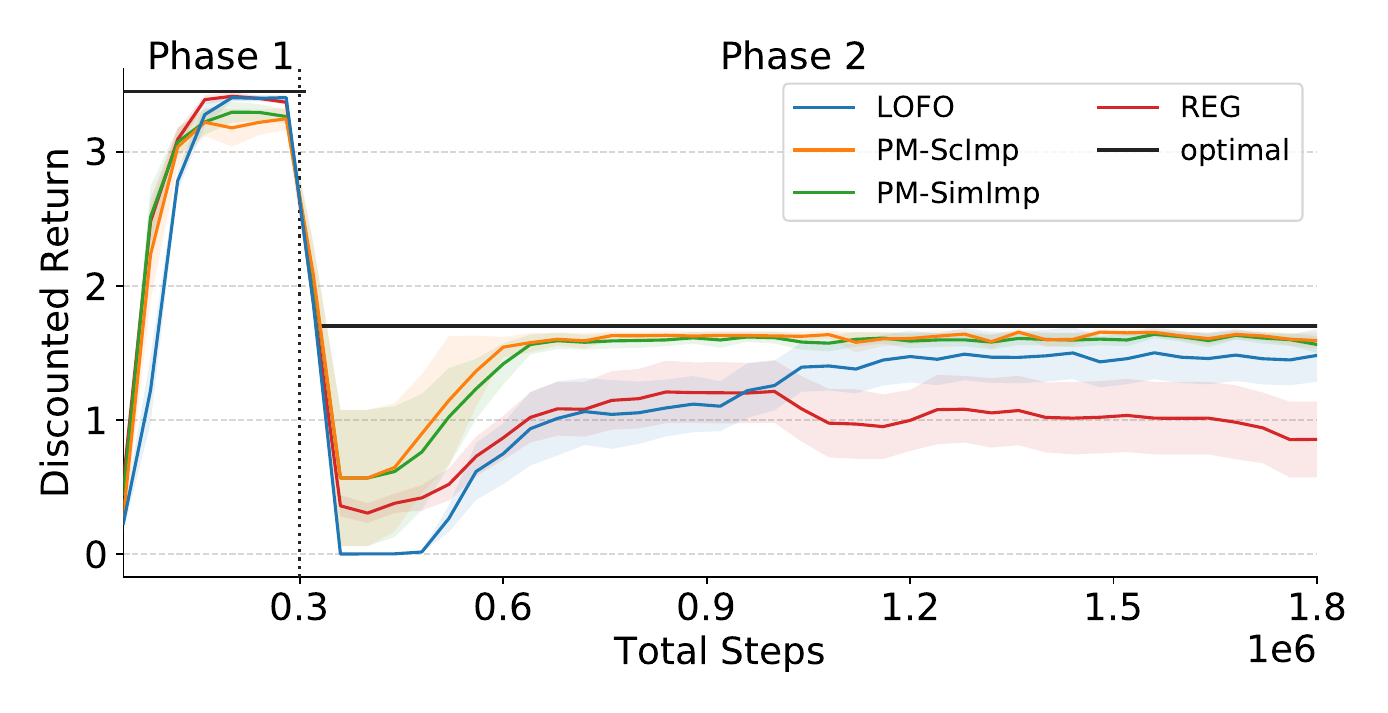}
        \caption{MiniGridLoCA} \label{fig:minigrid_plots_loca2}
    \end{subfigure}
    \caption{Plots showing the learning curves of the deep Dyna-Q agents that are referred to as PM-SimImp, PM-ScImp, REG and LOFO on the (a) MountainCarLoCA2 s and (b) MiniGridLoCA2 setups. Each learning curve is an average discounted return over 20 runs and the shaded area represents the confidence intervals. The maximum possible return in each phase is represented by a solid black line.}
    \label{fig:nonlinear_dyna_loca2}
\end{figure}

The evaluation results of the deep Dyna-Q agents on the LoCA1 and LoCA2 setups of the MountainCar and MiniGrid domains are presented in Fig.\ \ref{fig:nonlinear_dyna_loca1} \& \ref{fig:nonlinear_dyna_loca2}, respectively.

\subsection{PlaNet and Dreamer Experiments}
\label{app:add_loca_dreamer_planet_appendix}

\begin{figure}[]
    \centering
    \begin{subfigure}{0.31\textwidth}
        \centering
        \includegraphics[width=4.5cm]{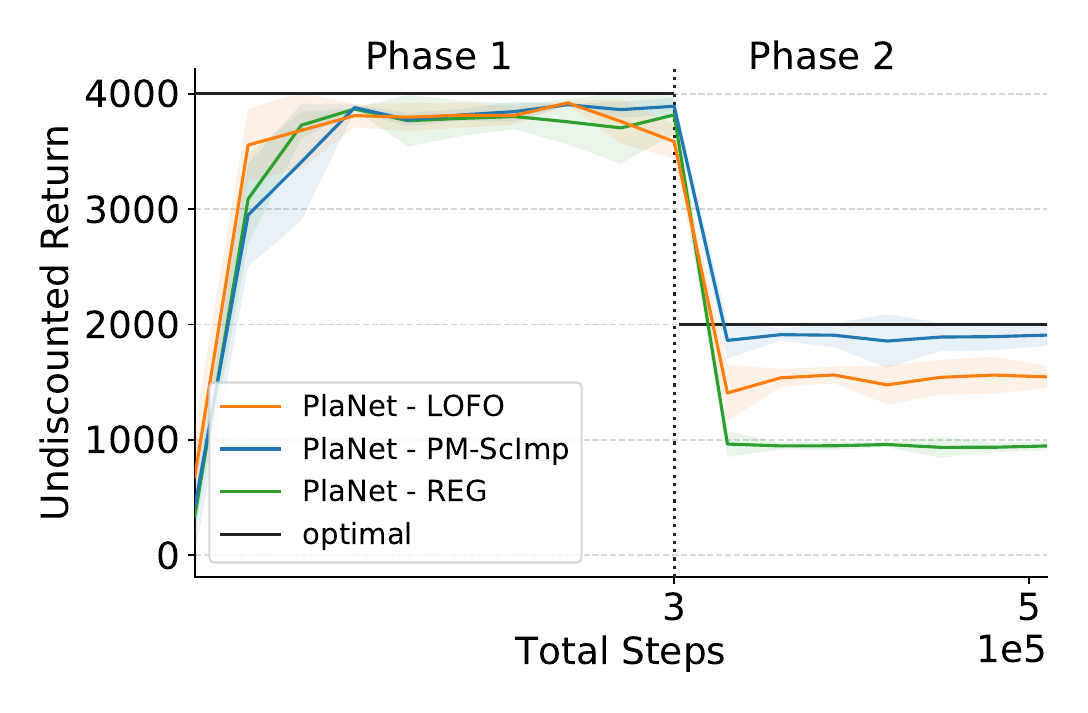}
        \caption{ReacherLoCA} \label{fig:reacherloca_planet_plots1}
    \end{subfigure}
    \begin{subfigure}{0.31\textwidth}
        \centering
        \includegraphics[width=4.5cm]{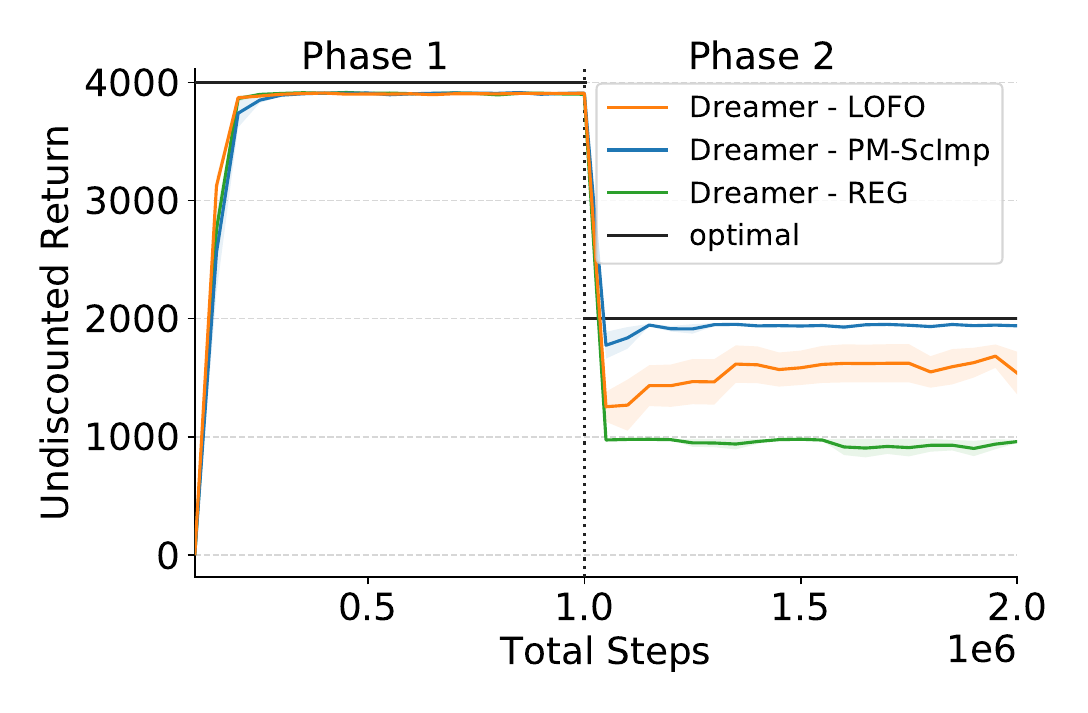}
        \caption{ReacherLoCA} \label{fig:reacherloca_dreamer_plots1}
    \end{subfigure}
    \begin{subfigure}{0.31\textwidth}
        \centering
        \includegraphics[width=4.5cm]{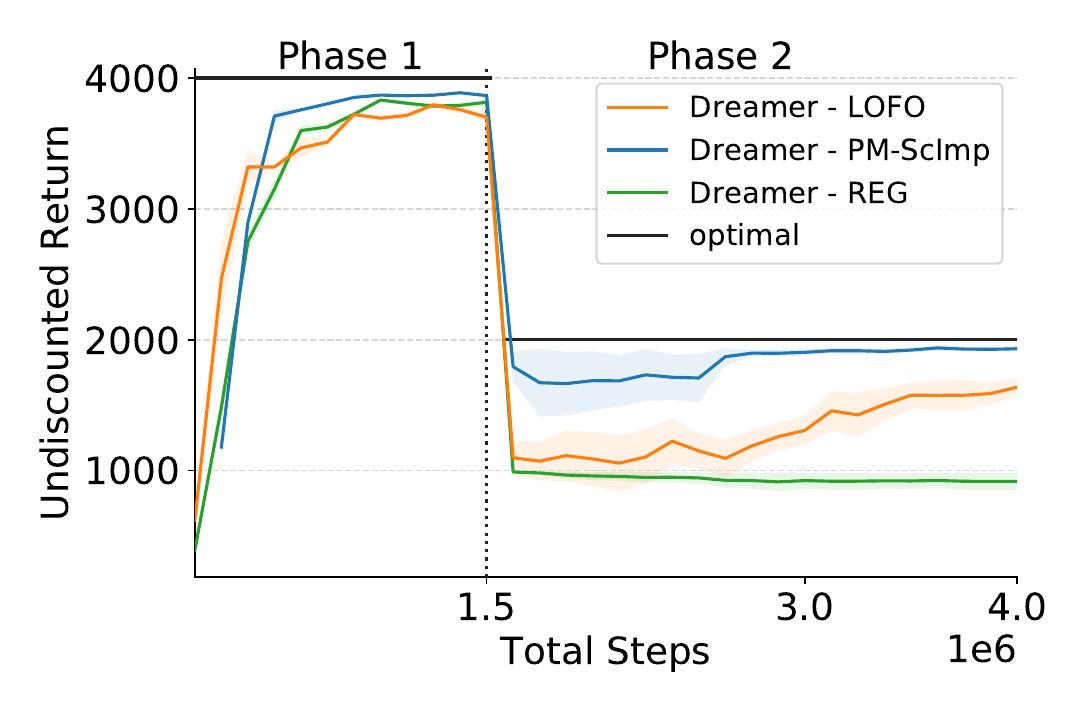}
        \caption{RandomReacherLoCA} \label{fig:random_reacherloca_dreamer_plots1}
    \end{subfigure}
    \caption{Plots showing the learning curves of the PlaNet - PM-ScImp, Dreamer - PM-ScImp, PlaNet - REG, Dreamer - REG, PlaNet - LOFO, Dreamer - LOFO agents on the (a, b) ReacherLoCA1 and (c) RandomReacherLoCA1 setups. Each learning curve is an average undiscounted return over 10 runs and the shaded area represents the confidence intervals. The maximum possible return in each phase is represented by a solid black line.}
    \label{fig:dreamer_and_planet1}
\end{figure}

\begin{figure}[]
    \centering
    \begin{subfigure}{0.31\textwidth}
        \centering
        \includegraphics[width=4.5cm]{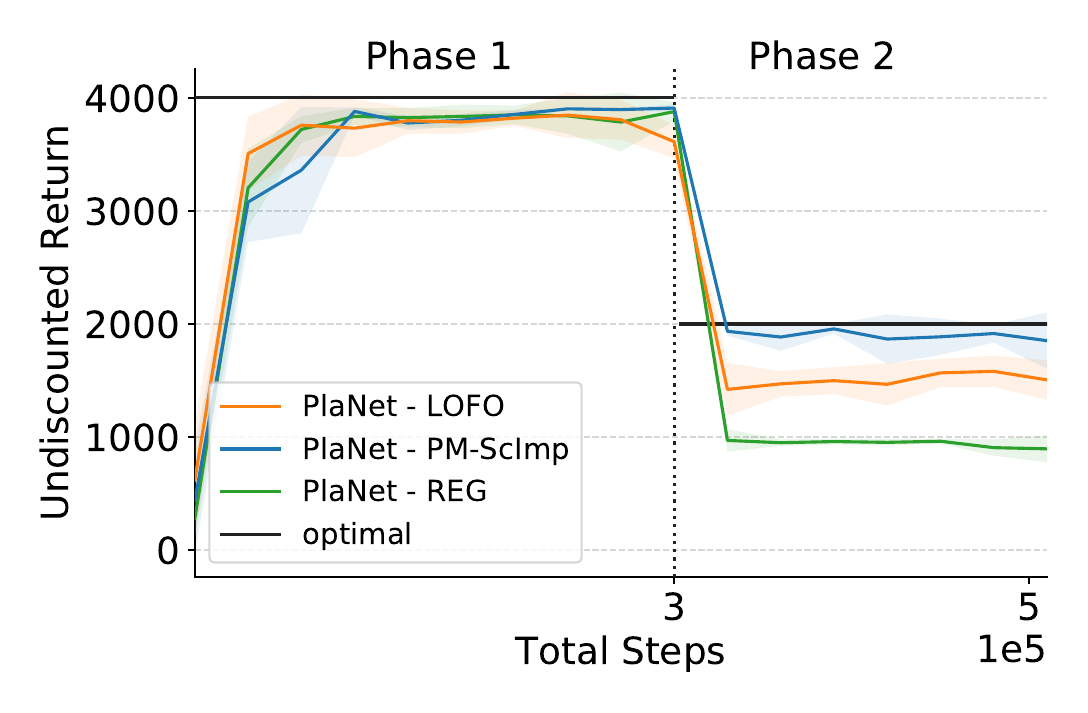}
        \caption{ReacherLoCA} \label{fig:reacherloca_planet_plots2}
    \end{subfigure}
    \begin{subfigure}{0.31\textwidth}
        \centering
        \includegraphics[width=4.5cm]{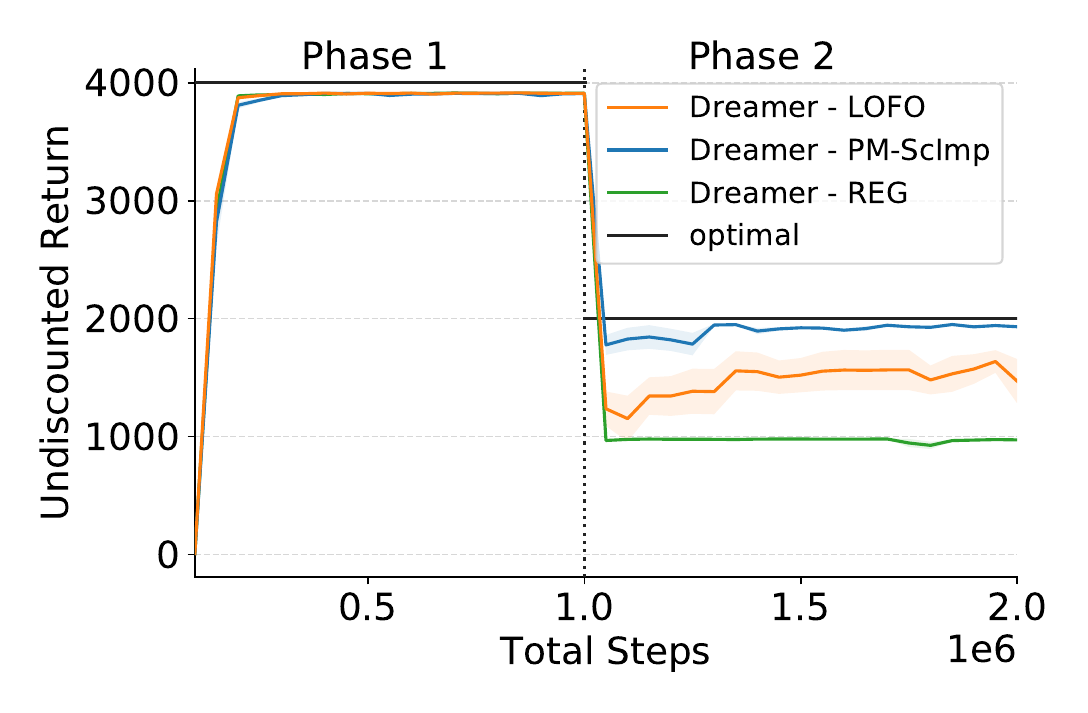}
        \caption{ReacherLoCA} \label{fig:reacherloca_dreamer_plots2}
    \end{subfigure}
    \begin{subfigure}{0.31\textwidth}
        \centering
        \includegraphics[width=4.5cm]{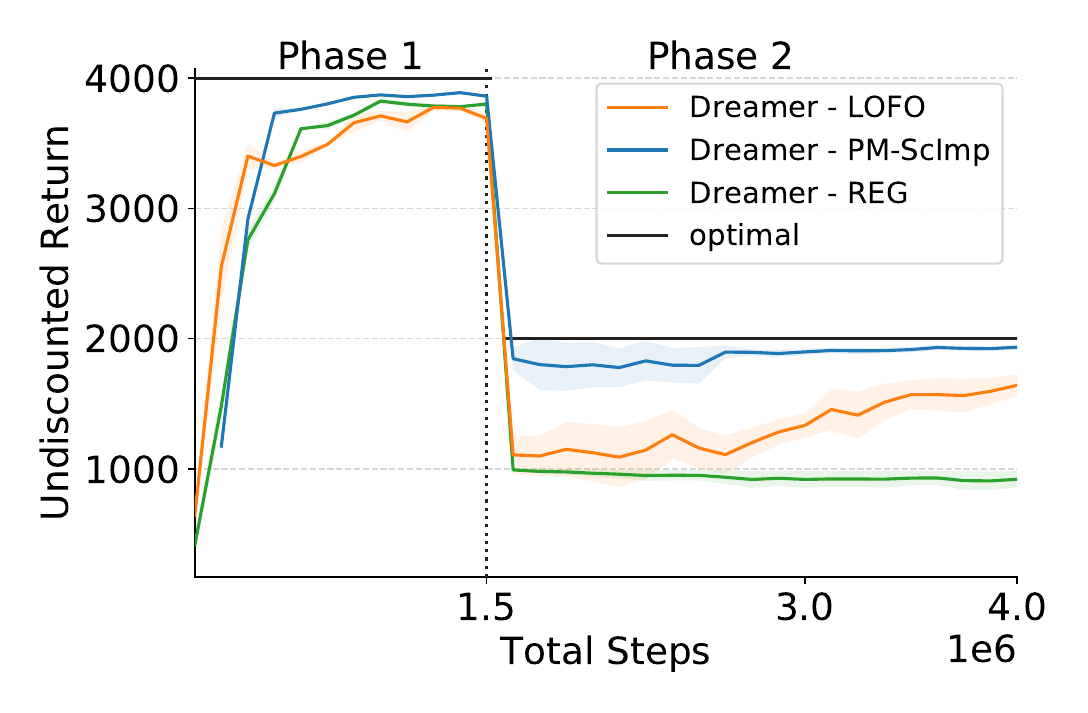}
        \caption{RandomReacherLoCA} \label{fig:random_reacherloca_dreamer_plots2}
    \end{subfigure}
    \caption{Plots showing the learning curves of the PlaNet - PM-ScImp, Dreamer - PM-ScImp, PlaNet - REG, Dreamer - REG, PlaNet - LOFO, Dreamer - LOFO agents on the (a, b) ReacherLoCA2 and (c) RandomReacherLoCA2 setups. Each learning curve is an average undiscounted return over 10 runs and the shaded area represents the confidence intervals. The maximum possible return in each phase is represented by a solid black line.}
    \label{fig:dreamer_and_planet2}
\end{figure}

The evaluation results of the PlaNet and Dreamer agents on the LoCA1 and LoCA2 setups of the Reacher and RandomReacher domains are presented in Fig.\ \ref{fig:dreamer_and_planet1} \& \ref{fig:dreamer_and_planet2}, respectively.

\end{document}